\documentclass{article}

\usepackage{microtype}
\usepackage{graphicx}
\usepackage{caption}
\usepackage{subcaption}
\usepackage{enumitem}
\usepackage{booktabs} %
\usepackage[utf8]{inputenc}
\usepackage{amssymb}
\usepackage{amsmath}
\usepackage{comment}
\usepackage{tabularx}
\usepackage[nodisplayskipstretch]{setspace}
\usepackage{amsthm}
\usepackage{mleftright}
\usepackage[normalem]{ulem}
\usepackage{makecell}
\usepackage{array}
\usepackage{import}

\usepackage{hyperref}

\usepackage{cleveref}

\newcolumntype{R}{>{\raggedleft\arraybackslash}X}
\newcolumntype{L}{>{\raggedright\arraybackslash}X}
\newcolumntype{C}{>{\centering\arraybackslash}X}
\newcolumntype{?}{!{\vrule width 1.5pt}}

\usepackage{xcolor}
\definecolor{darkgreen}{rgb}{0,0.5,0}

\DeclareMathOperator{\diag}{diag}

\newcommand{\tbf}{\fontseries{b}\selectfont}  %
\newcommand{\mccell}[1]{{\makecell[r]{#1}}}

\usepackage[accepted]{icml2021}

\icmltitlerunning{Relative Positional Encoding for Transformers with Linear Complexity}

\begin{document}

\twocolumn[
\icmltitle{Relative Positional Encoding for Transformers with Linear Complexity}

\icmlsetsymbol{equal}{*}

\begin{icmlauthorlist}
\icmlauthor{Antoine Liutkus}{equal,inria_mtp}
\icmlauthor{Ond\v{r}ej C\'ifka}{equal,imt}
\icmlauthor{Shih-Lun Wu}{rcit,ntu,ailabs}
\icmlauthor{Umut \c{S}im\c{s}ekli}{inria_paris}
\icmlauthor{Yi-Hsuan Yang}{rcit,ailabs}
\icmlauthor{Ga\"el Richard}{imt}
\end{icmlauthorlist}

\icmlaffiliation{inria_mtp}{Inria, Zenith Team, UMR LIRMM, Univ. Montpellier, France}
\icmlaffiliation{inria_paris}{INRIA~-- D\'{e}partement d'Informatique de l'\'{E}cole Normale Sup\'{e}rieure~-- PSL Research University, Paris, France}
\icmlaffiliation{imt}{LTCI, T\'el\'ecom Paris, Institut Polytechnique de Paris, France}
\icmlaffiliation{rcit}{Research Center for IT Innovation, Academia Sinica, Taiwan}
\icmlaffiliation{ntu}{National Taiwan University, Taiwan}
\icmlaffiliation{ailabs}{Taiwan AI Labs, Taiwan}

\icmlcorrespondingauthor{Liutkus Antoine}{antoine.liutkus@inria.fr}

\icmlkeywords{positional encoding, transformer, machine learning, ICML}

\vskip 0.25in
]

\printAffiliationsAndNotice{\icmlEqualContribution} %

\ifdefined\isaccepted%
\newcommand{\anon}[1]{{#1}}
\else%
\newcommand{\anon}[1]{[anonymized]}%
\fi

\begin{abstract}
Recent advances in Transformer models allow for unprecedented sequence lengths, due to linear space and time complexity. In the meantime, relative positional encoding (RPE) was proposed as beneficial for classical Transformers and consists in exploiting lags instead of absolute positions for inference. Still, RPE is not available for the recent linear-variants of the Transformer, because it requires the explicit computation of the attention matrix, which is precisely what is avoided by such methods. In this paper, we bridge this gap and present \textit{Stochastic Positional Encoding} as a way to generate PE that can be used as a replacement to the classical additive (sinusoidal) PE and provably behaves like RPE. The main theoretical contribution is to make a connection between positional encoding and cross-covariance structures of correlated Gaussian processes. We illustrate the performance of our approach on the Long-Range Arena benchmark and on music generation. 
\end{abstract}

\section{Introduction}
\label{introduction}

\subsection{Linear Complexity Transformers}

The Transformer model \cite{vaswaniAttentionAllYou} is a new kind of neural network that quickly became state-of-the-art in many application domains, including the processing of natural language \cite{heDeBERTaDecodingenhancedBERT2020}, images \cite{dosovitskiy2020image}, audio \cite{huangMusicTransformer2018,phamRelativePositionalEncoding2020} or bioinformatics \cite{alquraishi2019alphafold} to mention just a few. 

The core, novel component of the Transformer is the \textit{attention layer}. It computes $M$ output values $\textbf{y}_m$  from $N$ input values $\textbf{v}_n$, all being vectors of an arbitrary dimension. Following  classical non-parametric regression principles~\cite{nadaraya1964estimating, watson1964smooth}, it consists in a simple weighted sum:
\vspace{-0.1cm}
\begin{equation}
\textbf{y}_m=\frac{\sum_n a_{mn}\textbf{v}_n}{\sum_n a_{mn}}\,,\label{eq:output_with_attention}   
\vspace{-0.1cm}
\end{equation}
where each \textit{attention} coefficient $a_{mn}\in\mathbb{R_+}$ 
-- gathered in the $M\times N$ matrix $\textbf{A}$ -- indicates how important the value $\textbf{v}_n$ is in the computation of the output $\textbf{y}_m$.

One of the main %
contributions of the Transformer is %
an original method to compute %
these coefficients.
$D$-dimensional feature vectors $\textbf{k}_n$ and $\textbf{q}_m$ are attached to all items of the input and output sequences and are called \textit{keys} and \textit{queries}, respectively. Gathering them in the $N\times D$ and $M\times D$ matrices $\textbf{K}$ and $\textbf{Q}$, we get \textit{softmax dot-product attention} as: %
\vspace{-0.1cm}
\begin{equation}
\textbf{A}=\exp\mleft(\textbf{Q}\textbf{K}^{\top}\middle/\sqrt{D}\mright)
\equiv\left[a_{mn}=
\mathcal{K}\mleft(\textbf{q}_{m},\textbf{k}_{n}\mright)\right]_{mn},\label{eq:attention_kernel}\vspace{-0.1cm}
\end{equation}
where the function $\exp$ is applied element-wise. 
The right-hand side in (\ref{eq:attention_kernel}) is a generalization introduced by \citet{tsai2019transformer} and \citet{choromanskiRethinkingAttentionPerformers2020a}, where $\mathcal{K}$
is a \textit{kernel} function.
Parameters pertain to how keys $\textbf{k}_n$, values $\textbf{v}_n$ and queries $\textbf{q}_m$ are obtained from the raw sequences, usually by time-distributed fully connected layers.

\begin{figure}
     \centering
    \includegraphics[width=0.32\columnwidth]{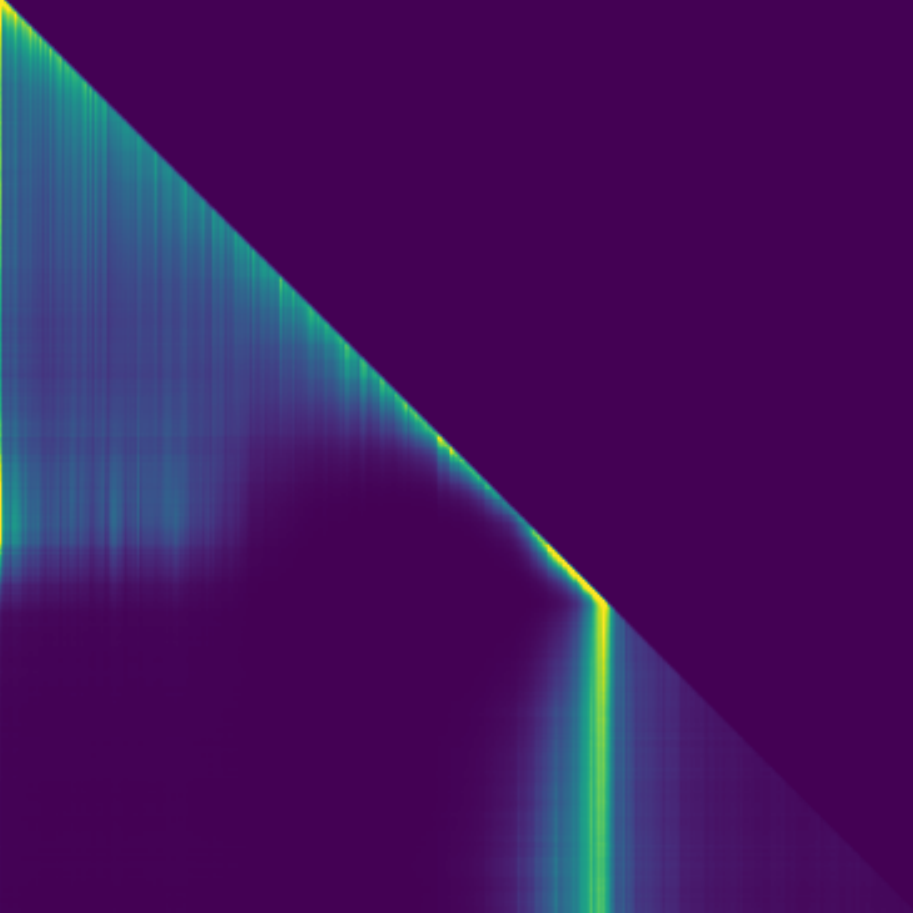}     \includegraphics[width=0.32\columnwidth]{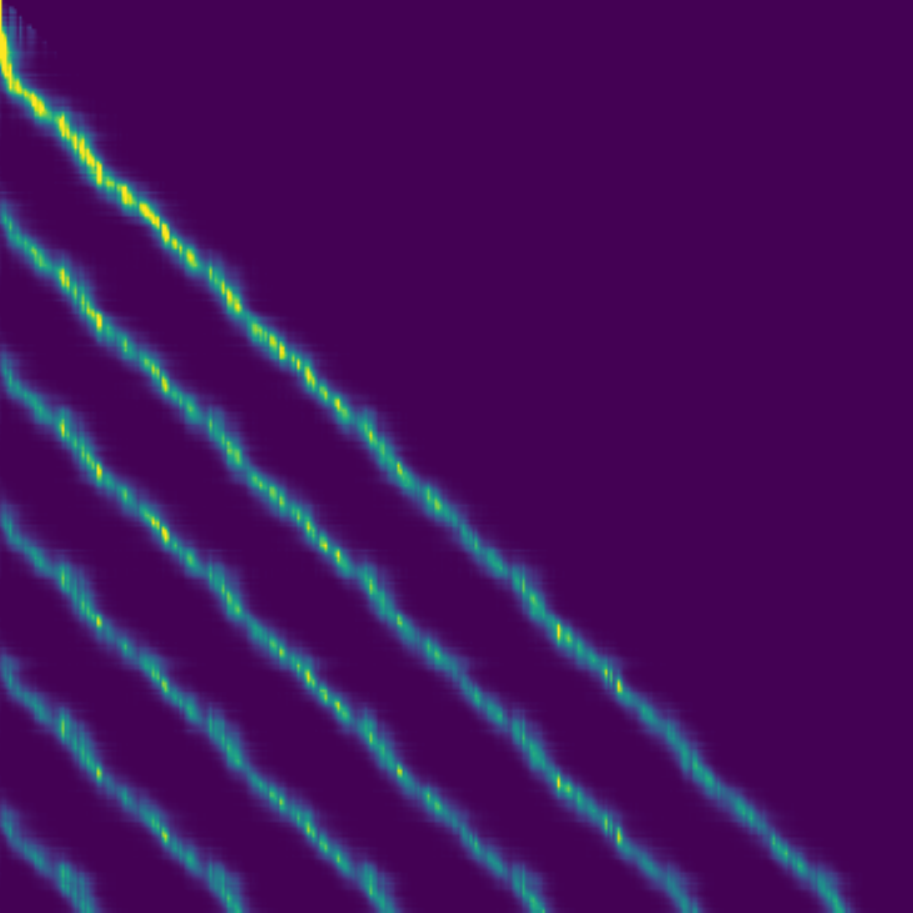} \includegraphics[width=0.32\columnwidth]{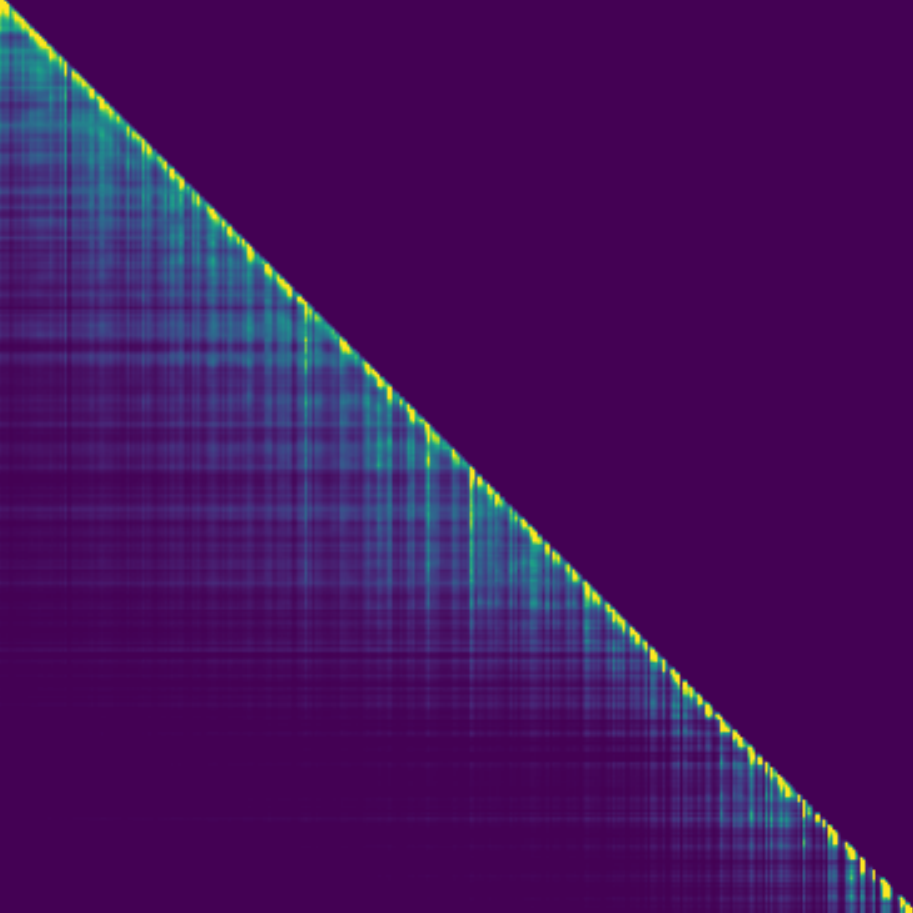}
    \vspace{-0.2cm}
\caption{Examples of attention patterns observed in the Performers trained for pop piano music generation (section~\ref{sec:unconditional-pop}) at inference time, for sequence length $M=N=3\,072$ while training sequences have length $2\,048$. (left) Absolute PE. (middle) Sinusoidal SPE. (right) Convolutional SPE. Note that SPE never requires computing these full attention patterns.\label{fig:attention_illustration}}
\end{figure}

The original Transformer architecture \cite{vaswaniAttentionAllYou} explicitly computes the attention matrix $\textbf{A}$, leading to a $\mathcal{O}\mleft(MN\mright)$ complexity that prevents it from scaling to very long sequence lengths. Although this is not necessarily a problem when sequence lengths are barely on the order of a few hundreds, as in some language processing tasks, it is prohibitive for very large signals like high-resolution images or audio.

Focusing on this scalability issue, several approaches have been recently investigated to allow for long sequences:
\vspace{-0.5em}
\begin{itemize}[leftmargin=*, noitemsep,topsep=0pt]
    \item \textit{Attention clustering} schemes  group items among which dependencies are computed through regular attention. This is either done by using simple proximity rules within the sequences, leading to chunking strategies \cite{daiTransformerXLAttentiveLanguage2019}, or by clustering the keys and values \cite{royEfficientContentBasedSparse2020}. Inter-cluster dependencies are either ignored or summarized via fixed-length context vectors that are coined in as \textit{memory} \cite{wuMemformerMemoryAugmentedTransformer2020}.
    \item Assuming the attention matrix to be \textit{sparse}. In this case, only a few $a_{mn}$ are nonzero \cite{childGeneratingLongSequences2019}.
    \item Assuming $\textbf{A}$ has 
    a particular (low-rank) \textit{structure} and can be decomposed as the product of two smaller matrices. A prototypical example is the Linformer~\cite{wangLinformerSelfAttentionLinear2020}, which is limited to fixed-length inputs. Another very recent line of research in this same vein takes:
    \vspace{-0.2cm}
\begin{equation}
        \textbf{A}\approx\phi\mleft(\textbf{Q}\mright)\phi\mleft(\textbf{K}\mright)^\top,\label{eq:features_linear_attention}
\vspace{-0.2cm}
\end{equation}    
where $\phi:\mathbb{R}^D\rightarrow\mathbb{R}^R$ is a non-linear \textit{feature map} applied to each key $\textbf{k}_n$ and query $\textbf{q}_m$, and $R\ll\min(M,N)$ \cite{shenEfficientAttentionAttention2020, katharopoulosTransformersAreRNNs}.
    \item  When $\mathcal{K}$ in (\ref{eq:attention_kernel}) is a positive (semi)definite kernel, the Performer \cite{choromanskiRethinkingAttentionPerformers2020a} leverages \textit{reproducing kernel Hilbert spaces} to show that a random $\phi$ may be used to exploit this convenient decomposition~(\ref{eq:features_linear_attention}) \textit{on average}, even when $\textbf{A}$ is not low rank:
\vspace{-0.2cm}
    \begin{equation}
\mathcal{K}\succeq 0 \Leftrightarrow \textbf{A}=\mathbb{E}_\phi\left[\phi\mleft(\textbf{Q}\mright)\phi\mleft(\textbf{K}\mright)^\top\right],\label{eq:FAVOR} 
    \vspace{-0.2cm}
    \end{equation}
    where $\phi$ is drawn from a distribution that depends on $\mathcal{K}$. 
    A simple example is 
    $\phi_\textbf{W}(\textbf{k}_n)=\max(0,\textbf{W}\textbf{k}_n)$, 
    with a random $\textbf{W}\in\mathbb{R}^{R\times D}$ for some $R\in\mathbb{N}$.
\end{itemize}
\vspace{-2pt}
Whenever an efficient scheme like~(\ref{eq:features_linear_attention}) or~(\ref{eq:FAVOR}) is used, the outputs can be obtained without
computing
the attention coefficients~$a_{mn}$, as in (\ref{eq:compute_output_without_full_attention}).\footnote{A somewhat related strategy is used by the recent LambdaNetworks \cite{bello2020lambdanetworks}, which encapsulate the key-value information as a so-called \textit{lambda} function to be applied query-wise, hence also avoiding the computation of a full attention matrix.}
\vspace{-0.1cm}

\subsection{Positional Encoding}
In Transformer networks, the outputs $\textbf{y}_m$ are computed as linear combinations of \textit{all} input values $\textbf{v}_n$, weighted by attention coefficients $a_{mn}$. In sequence modeling, it is reasonable to assume that the actual \textit{positions} $m$ and $n$ should play a role in the computation, in addition to the \textit{content} at these locations; otherwise, any permutation of the sequence would lead to the same output.
Two core approaches were undertaken to incorporate position information:
\vspace{-0.5em}
\begin{itemize}[leftmargin=*, noitemsep,topsep=0pt]
\item The original Transformer \cite{vaswaniAttentionAllYou} adds %
this information to the inputs of the network, i.e.\ before the first attention layer. This can be equivalently understood as augmenting the keys, values and queries:
\begin{equation}
\arraycolsep=1.5pt
\begin{array}{ccc}
\textbf{k}_{n}\leftarrow\textbf{k}_{n}+\overline{\textbf{k}}_{n}, & \textbf{v}_{n}\leftarrow\textbf{v}_{n}+\overline{\textbf{v}}_{n}, & \textbf{q}_{m}\leftarrow\textbf{q}_{m}+\overline{\textbf{q}}_{m},\end{array} \label{eq:additive_PE_keys_domain}
\end{equation}
where we write $\overline{\textbf{k}}_n\in\mathbb{R}^D$ for the \textit{keys positional encoding} (PE; \citealp{sukhbaatarEndToEndMemoryNetworks2015}) at position $n\in\mathbb{N}$
and analogously for values and queries. 
\citeauthor{vaswaniAttentionAllYou}\ propose a deterministic scheme based on trigonometric functions, which is shown to work as well as trainable embeddings. 
\item As an example of positional encoding \textit{in the attention domain}, a \textit{relative positional} encoding (RPE) was proposed by \citet{shawSelfAttentionRelativePosition2018}, building on the idea that time lags $m-n$ are more important than absolute positional encoding (APE) for prediction.
It is written as:
\vspace{-0.5cm}
\begin{align}
\textbf{A} & =\exp\mleft(\mleft(\textbf{Q}\textbf{K}^{\top}+\boldsymbol{\Omega}\mright)\middle/\sqrt{D}\mright),\text{ with:}\\
\boldsymbol{\Omega} & \equiv\left[\omega_{mn}=\sum_{d=1}^{D}q_{md}\mathcal{P}_{d}\mleft(m-n\mright)\right]_{mn}.
\vspace{-0.2cm}
\label{eq:original_RPE}
\end{align}
The terms $\mathcal{P}_d$ now act as $D$ different encodings for \textit{time lags} selected based on the queries.
This change is advocated as bringing important performance gains in many application areas and has enjoyed a widespread use ever since.%
\end{itemize}

Although writing down the positional encoding in the attention domain is beneficial for performance~\cite{shawSelfAttentionRelativePosition2018,daiTransformerXLAttentiveLanguage2019,tsai2019transformer}, we are only aware of implementations that either require the computation of $\textbf{A}$, or clustered attention schemes, which \textit{in fine}
decompose $\textbf{A}$ into smaller attention matrices, and  \textit{compute them}. %
This is in sharp contrast to  (\ref{eq:features_linear_attention}) and (\ref{eq:FAVOR}), which never compute the attention matrix.

\textbf{Our contributions} can be summarized as follows:
\vspace{-0.5em}\begin{itemize}[leftmargin=*, noitemsep,topsep=0pt]

\item We propose \textit{Stochastic Positional Encoding} (SPE) as a general PE scheme
 \textit{in the keys domain}, that enforces a particular attention pattern devised \textit{in the attention domain}. This enables RPE without explicit computation of attention. To our knowledge, it is the first RPE strategy that is compatible with $\mathcal{O}\mleft(N\mright)$  Transformers like \citet{choromanskiRethinkingAttentionPerformers2020a} and \citet{katharopoulosTransformersAreRNNs}.
\item We study the impact of SPE on performance on the Long-Range Arena benchmark \cite{tay2021long} and two music generation tasks.
Since RPE was so far limited to short sequences, we believe this is the first study of its advantages on long-range predictions. Our results demonstrate better validation losses and extrapolation ability.
\item We provide additional resources on our companion website,\footnote{\url{https://cifkao.github.io/spe/}} including Python implementations of SPE for PyTorch and JAX/Flax.

\end{itemize}

\section{Stochastic Positional Encoding}
\textbf{Index set and notation}. We assume that the input/output sequences are indexed by $n,m\in\mathbb{T}$, where $\mathbb{T}$ is the \textit{index set}. For regularly sampled sequences, we have $\mathbb{T}=\mathbb{N}$, but more settings are possible, like irregularly sampled time series ($\mathbb{T}=\mathbb{R}$) or images ($\mathbb{T}=\mathbb{N}^2)$. In any case, the particular lists of input~/~output locations under consideration are written: $\mathcal{N}$ and $\mathcal{M}$, with respective sizes $N$ and $M$ (the case $\mathcal{N}=\mathcal{M}$ is called \emph{self-attention}). The corresponding keys and values are
hence indexed as $\left\{\textbf{k}_n\right\}_{n\in\mathcal{N}}$ and $\left\{\textbf{v}_n\right\}_{n\in\mathcal{N}}$, while queries are $\left\{\textbf{q}_m\right\}_{m\in\mathcal{M}}$. For convenience, we write $a_{mn}$ for the entries of the $M\times N$ attention matrix $\textbf{A}$.
\\We use bold uppercase for matrices, bold lowercase for vectors and a NumPy-like notation: if $\textbf{X}_k$ is a $I\times J$ matrix, $\textbf{x}_{k,i}$ and $\textbf{x}_{k,:,j}$ stand for its $i^{th}$ row and $j^{th}$ column, respectively. 

\textbf{Assumptions.} In the remainder of this paper, we will seek an attention matrix $\textbf{A}$ given by:
\vspace{-0.1cm}
\begin{equation}
\textbf{A}=\exp\mleft(\left[\sum_{d=1}^{D}q_{md}\mathcal{P}_{d}\mleft(m,n\mright)k_{nd}\right]_{mn}\middle/\sqrt{D}\mright),\label{eq:PE_in_attention_domain}
\vspace{-0.1cm}\end{equation}
where $\{\mathcal{P}_d\}_{d=1}^D$ are \textit{position kernels}. Defining $\textbf{P}_d\equiv\mleft[\mathcal{P}_d(m,n)\mright]_{mn}$, this can be written in matrix form as:
\vspace{-0.1cm}
\begin{equation}
\textbf{A}=\exp\mleft(\sum_{d=1}^{D}\mleft.\diag\mleft(\textbf{q}_{:,d}\mright)\textbf{P}_{d}\diag\mleft(\textbf{k}_{:,d}\mright)\middle/\sqrt{D}\mright.\mright),\label{eq:PE_in_attention_domain_matrix_form}
\vspace{-0.15cm}\end{equation}
which is understood as having $D$ positional attention templates $\textbf{P}_d$ jointly activated by the queries $\textbf{q}_{:,d}$ and keys $\textbf{k}_{:,d}$. Original RPE (\ref{eq:original_RPE}) can be seen as a special case, where some entries are kept constant.

\begin{algorithm} [t]
\textbf{Input}
\begin{itemize}[leftmargin=*, noitemsep,topsep=0pt]
    \item position kernel $\mathcal{P}\mleft(m,n\mright)$, number of replicas $R$.
    \item initial $M\times D$ and $N\times D$ queries $\textbf{Q}$ and keys $\textbf{K}$.
\end{itemize}
\textbf{Positional encoding:}
\begin{itemize}[leftmargin=*, noitemsep,topsep=0pt]
    \item Draw the $D$ independent couples $\{\overline{\textbf{Q}}_d, \overline{\textbf{K}}_d\}_d$ of $M\times R$ and $N\times R$ matrices  as in section~\ref{ss:drawing_gp}
    \item Set $\widehat{\textbf{Q}}$ and $\widehat{\textbf{K}}$ as in (\ref{eq:SPE_PE_Q}) and (\ref{eq:SPE_PE_K})
\end{itemize}

\vspace{0.05cm}
\textbf{Inference} compute outputs $\textbf{Y}$ with  the $\mathcal{O}(N)$ Transformer:\vspace{-5pt}
\begin{equation}
\textbf{Y}\leftarrow\text{diag}\mleft(\textbf{d}\mright)^{-1}\left[\phi\mleft(\widehat{\textbf{Q}}\mright)\left[\phi\mleft(\widehat{\textbf{K}}\mright)^\top\textbf{V}\right]\right]\label{eq:compute_output_without_full_attention}
\vspace{-2pt}
\end{equation}
with $\textbf{d}=\phi(\widehat{\textbf{Q}})\mleft[\phi\mleft(\widehat{\textbf{K}}\mright)^\top\textbf{1}_{N}\mright]$ and $\phi$ discussed in (\ref{eq:features_linear_attention})/(\ref{eq:FAVOR}).

\caption{Stochastic Positional Encoding.\label{alg:SPE}}
\end{algorithm}

\textbf{Positional attention as covariance.} The key idea for SPE is to see the attention kernel $\mathcal{P}_d(m,n)$ as a \textit{covariance}: 
\begin{equation}
    \left(\forall\mathcal{M},\mathcal{N}\right)\left(\forall m,n\right)\mathcal{P}_d\mleft(m,n\mright)=\mathbb{E}\left[\overline{Q}_{d}(m)\overline{K}_{d}(n)\right],
\label{eq:cross-covariance}\end{equation}
where $\overline{Q}_d(m)$ and $\overline{K}_d(n)$ are two real and zero-mean random variables, which will be chosen with the single condition that their covariance function matches $\mathcal{P}_d$. Semantically, they should be understood as (randomly) encoding position $m$ for queries and position $n$ for keys, respectively. When multiplied together as in dot-product attention, they yield the desired attention template $\mathcal{P}_d(m,n)$ on average. The central intuition is that the actual positional encodings do not matter as much as their dot-product.

In what follows, we will impose specific structures on the cross-covariance  $\mathcal{P}_d(m,n)$, which will in turn allow us to design \emph{random processes} $\overline{Q}_d = \{\overline{Q}_d(m)\}_{m\in \mathcal{M}}$ and $\overline{K}_d = \{\overline{K}_d(n)\}_{n \in \mathcal{N}}$ such that \eqref{eq:cross-covariance} holds.
The core advantage of this construction is to allow for $\textbf{P}_d$ to be factorized. Let us for now assume that we construct the distributions of $\{\overline{Q}_d(m),\overline{K}_d(n)\}_d$ in such a way that we can sample from them
(we will see how in section~\ref{ss:drawing_gp}) and consider $R$ independent realizations of them for given $\mathcal{M}$ and $\mathcal{N}$, gathered in the $M\times R$ and $N\times R$ matrices $\overline{\textbf{Q}}_d$ and $\overline{\textbf{K}}_d$:
\begin{equation}
\overline{\textbf{Q}}_d\equiv[q_{d,m,r}\sim \overline{Q}_d(m)]_{mr},  ~~\overline{\textbf{K}}_d\equiv[k_{d,n,r}\sim \overline{K}_d(n)]_{nr}.
\end{equation}

For large $R$, by the law of large numbers, we obtain:
\vspace{-0.05cm}\begin{equation}
    \textbf{P}_d\approx\mleft[\overline{\textbf{Q}}_d  \overline{\textbf{K}}_d^\top\mright] /R .\label{eq:Pd_equal_QK}
\vspace{-0.1cm}\end{equation}
This leads $\textbf{A}$ in (\ref{eq:PE_in_attention_domain_matrix_form}) to be given by: %
\vspace{-0.05cm}
\begin{align}
\textbf{A} & \approx\exp\mleft(\sum_{d=1}^{D}\diag\mleft(\textbf{q}_{:,d}\mright)\frac{\overline{\textbf{Q}}_{d}\overline{\textbf{K}}_{d}^{\top}}{R}\diag\mleft(\textbf{k}_{:,d}\mright)/\sqrt{D}\mright)\label{eq:E_as_a_sum_of_product}\\
 & \approx\exp\frac{\mleft(\sum\limits_{d=1}^{D}\diag\mleft(\textbf{q}_{:,d}\mright)\overline{\textbf{Q}}_{d}\mright)\mleft(\sum\limits_{d=1}^{D}\diag\mleft(\textbf{k}_{:,d}\mright)\overline{\textbf{K}}_{d}\mright)^{\top}}{R\sqrt{D}}\label{eq:E_as_a_product}.
 \vspace{-0.25cm}
\end{align}
Here, a \textit{crucial} observation is that for large $R$, the cross-terms $\overline{\textbf{Q}}_d \overline{\textbf{K}}^\top_{d'\neq d}$  are negligible due to independence, provided that the means of the processes are selected to be zero. Finally, picking queries and keys as:
\vspace{-0.2cm}\begin{align}
    \widehat{\textbf{Q}} & \leftarrow\sum_{d=1}^{D}\diag\mleft(\textbf{q}_{:,d}\mright)\overline{\textbf{Q}}_{d}/\sqrt[4]{DR}\,,\label{eq:SPE_PE_Q}\\
\widehat{\textbf{K}} & \leftarrow\sum_{d=1}^{D}\diag\mleft(\textbf{k}_{:,d}\mright)\overline{\textbf{K}}_{d}/\sqrt[4]{DR}\label{eq:SPE_PE_K}\,,
\vspace{-0.25cm}\end{align}
we see from (\ref{eq:E_as_a_product}-\ref{eq:SPE_PE_K}) that we get back to the usual multiplicative scheme~(\ref{eq:attention_kernel})
with $\textbf{A}=\exp(\widehat{\textbf{Q}}\widehat{\textbf{K}}^\top/\sqrt{R})$, where the queries/keys now have dimension $R$ and can be used in (\ref{eq:compute_output_without_full_attention}) to directly get outputs without computing $\textbf{A}$.

The procedure is summarized in Algorithm~\ref{alg:SPE}: we provide a way (\ref{eq:SPE_PE_Q}-\ref{eq:SPE_PE_K}) to achieve PE in the \textit{keys domain}, such that the desired model (\ref{eq:PE_in_attention_domain}) is enforced in the \textit{attention domain}, parameterized by the attention kernels $\mathcal{P}_d$. Interestingly, this is done without ever computing attention matrices, complying with $\mathcal{O}(N)$ Transformers. 
The remaining challenge, which we discuss next, is to generate $\overline{\textbf{Q}}_d$ and $\overline{\textbf{K}}_d$ enforcing (\ref{eq:Pd_equal_QK}).

\subsection{Drawing Stochastic Positional Encodings}

\label{ss:drawing_gp}

\vspace{-0.1cm}Inspecting (\ref{eq:cross-covariance}), we notice that our objective is to draw samples from $D$ pairs of centered random processes $\left\{  \overline{Q}_d,\overline{K}_d\right\}_d$, with a prescribed cross-covariance structure $\mathcal{P}_d$. It is reasonable to use Gaussian processes for this purpose \cite{williamschristopherkiGaussianProcessesMachine2006}, which have the maximum entropy for known mean and covariance. %
Such distributions are frequently encountered in geophysics in the \textit{co-kriging} literature \cite{matheron1963principles, genton2015cross}, where scientists routinely handle correlated random fields. The particular twists of our setup are: we have a \textit{generative} problem, e.g.\ as in \citet{vorechovsky2008simulation}; however, as opposed to their setting, we are not directly interested in the marginal covariance function of each output, provided that the desired cross-covariance structure holds.

The most straightforward application of SPE arises when we pick $\mathcal{P}_d\mleft(m,n\mright)=\mathcal{P}_d\mleft(m-n\mright)$, i.e.\ a stationary  position kernel, which was coined in as choosing \textit{relative} attention in \citet{shawSelfAttentionRelativePosition2018} and boils down to enforcing a \emph{Toeplitz} structure 
for the cross-covariance matrix $\textbf{P}_d\equiv\left[\mathcal{P}_d\mleft(m-n\mright)\right]_{m,n}$ between $\overline{Q}_d$ and $\overline{K}_d$.

\begin{figure*}
     \centering
    \noindent\fbox{\begin{minipage}[t]{0.518\textwidth}%
    \includegraphics[width=1\textwidth]{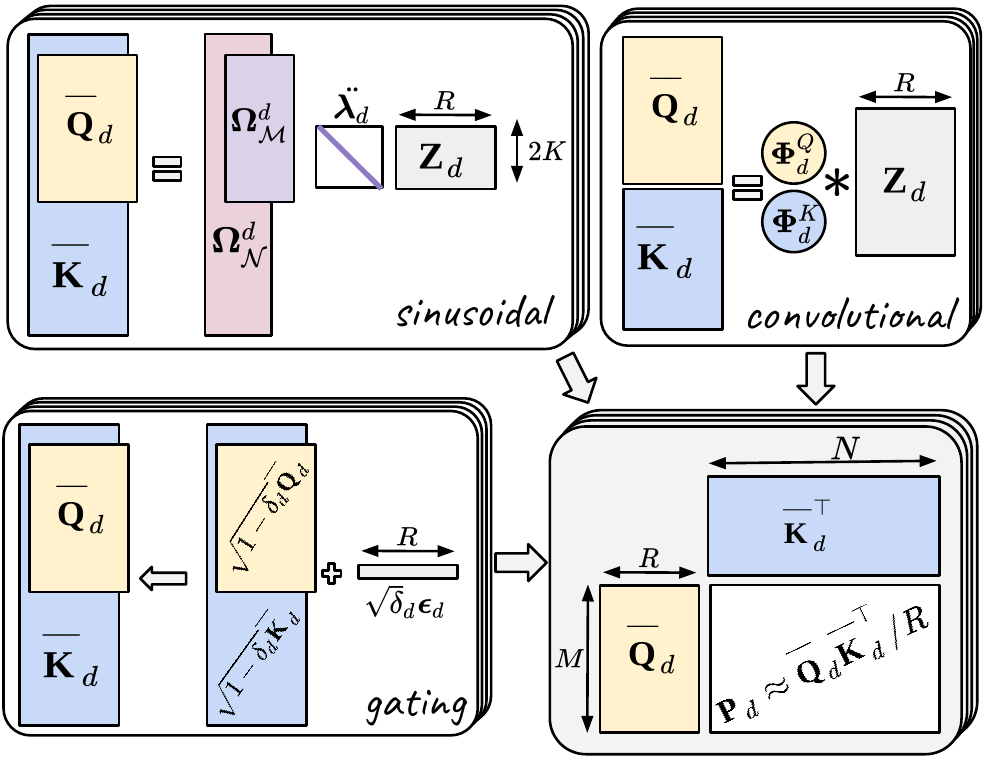}
    \end{minipage}}\hspace{0.2cm} \fbox{\begin{minipage}[t]{0.43\textwidth}%
    \includegraphics[width=1\textwidth]{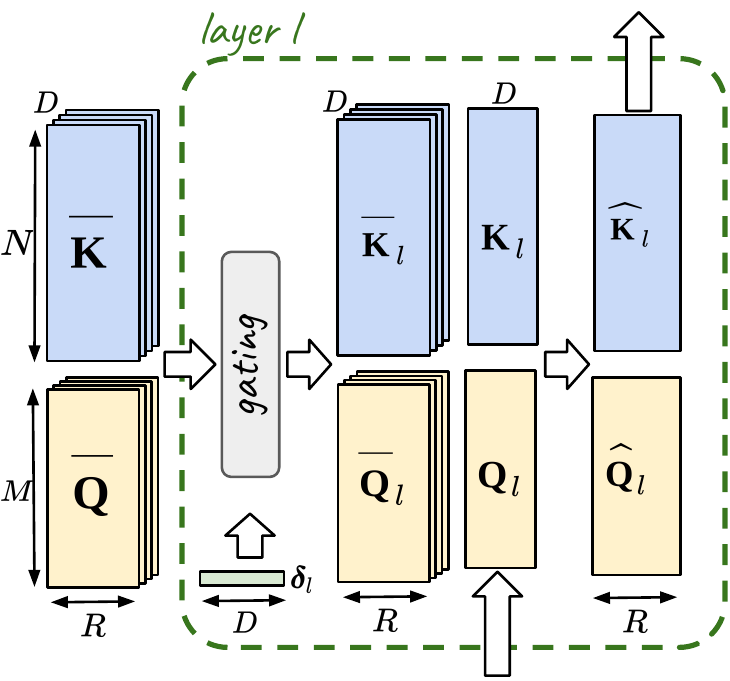}
    \end{minipage}}
\caption{(left) Generation of $\overline{\textbf{Q}}$ and $\overline{\textbf{K}}$ in SPE, which approximate the templates $\textbf{P}_d$ when multiplied together.
(right) $\overline{\textbf{Q}}$ and $\overline{\textbf{K}}$ can be shared across layers. At each layer~$l$, different gating is (optionally) used, before applying (\ref{eq:SPE_PE_Q}-\ref{eq:SPE_PE_K}) to generate new queries $\textbf{Q}$ and keys $\textbf{K}$.\label{fig:SPE_illustration}}
\end{figure*}

We propose two variants of SPE to handle this important special case, illustrated in Figure~\ref{fig:SPE_illustration}. The first variant yields \emph{periodic} covariance functions. It can be beneficial whenever attention should not vanish with large lags, as in traffic prediction \cite{xue2020trailer} or, as we show, in music generation. The second variant generates \emph{vanishing} covariance functions; a concept which has recently been shown useful \cite{wang2021on}, and notably yields smaller validation losses in some of our experiments.

\textbf{Variant I. Relative and periodic attention} (\texttt{sineSPE}). In our first approach, we consider the case where $\mathcal{P}_d$ is periodic,
which gets a convenient treatment.
We assume:
\vspace{-0.2cm}\begin{equation}
    \mathcal{P}_d(m,n)=\sum_{k=1}^{K} \lambda_{kd}^2 \cos\mleft(2\pi f_{kd}\left(m-n\right) + \theta_{kd}\mright)\,,\label{eq:periodic_pd}
\vspace{-0.2cm}
\end{equation}
where $K\in\mathbb{N}$ is the number of \emph{sinusoidal} components and $\textbf{f}_{d}\in[0~1]^K$, $\boldsymbol{\theta}_{d}\in[-\pi~\pi]^K$ and $\boldsymbol{\lambda}_d\in\mathbb{R}^K$ gather their $K$ frequencies, phases, and weights, respectively. 
By using the matrix notation, we can rewrite (\ref{eq:periodic_pd}) as:
\vspace{-0.15cm}
\begin{equation}
\textbf{P}_{d}=\boldsymbol{\Omega}\mleft(\mathcal{M},\boldsymbol{f}_{d},\boldsymbol{\theta}_{d}\mright)\diag\mleft(\ddot{\boldsymbol{\lambda}{}_{d}}\mright)^{2}\boldsymbol{\Omega}\mleft(\mathcal{N},\boldsymbol{f}_{d},\textbf{0}\mright)^{\top},\label{eq:decomposition_toeplitz}
\end{equation}
where $\ddot{\textbf{v}}\equiv\left[v_{\left\lfloor p/2\right\rfloor }\right]_{p}\in\mathbb{R}^{2K}$
denotes a twice upsampled version of a vector $\textbf{v}\in\mathbb{R}^{K}$, $\lfloor \cdot \rfloor$ denotes the floor operation, and for an index set $\mathcal{I}$, $\boldsymbol{\Omega}(\mathcal{I}, \textbf{a}, \textbf{b})$ is a matrix of size $\left|\mathcal{I}\right|\times 2K$, with entries (0-based indexing):
\vspace{-0.15cm}\[
\left[\boldsymbol{\Omega}\left(\mathcal{I},\boldsymbol{a},\boldsymbol{b}\right)\right]_{nl}=\begin{cases}
\cos\mleft(2\pi a_{k}n+b_{k}\mright) & \text{if }l=2k\\
\sin\mleft(2\pi a_{k}n+b_{k}\mright) & \text{if }l=2k+1
\end{cases}
\vspace{-0.2cm}\]

It can be shown that if $\boldsymbol{\theta}_{d}=\textbf{0}$ and $\mathcal{M}=\mathcal{N}$, we get back to the (unique) Vandermonde decomposition for positive definite Toeplitz matrices\footnote{If $\textbf{P}_d\succeq 0$ and $K\geq N$, (\ref{eq:decomposition_toeplitz}) still holds but is not unique.} \cite{yang2016vandermonde}, which boils down in our context to assuming that $\forall\tau,\mathcal{P}_d(0)\geq\mathcal{P}_d(\tau)$. Since this is not always desirable, we keep the more general~(\ref{eq:decomposition_toeplitz}).

At this point, we can easily build $\overline{\textbf{Q}}_d$ and $\overline{\textbf{K}}_d$. We draw a $2K\times R$ matrix $\textbf{Z}_d$ with independent and identically distributed (i.i.d.) Gaussian entries of unit variance, and define:
\begin{align}
\overline{\textbf{Q}}_{d}&\leftarrow\boldsymbol{\Omega}\mleft(\mathcal{M},\boldsymbol{f}_{d},\boldsymbol{\theta}_{d}\mright)\diag\mleft(\ddot{\boldsymbol{\lambda}{}_{d}}\mright)\textbf{Z}_{d}/\sqrt{2K}\label{eq:toeplitz_Qbar}\,,\\
\overline{\textbf{K}}_{d}&\leftarrow\boldsymbol{\Omega}\mleft(\mathcal{N},\boldsymbol{f}_{d},\textbf{0}\mright)\diag\mleft(\ddot{\boldsymbol{\lambda}{}_{d}}\mright)\textbf{Z}_{d}/\sqrt{2K}\,.\label{eq:toeplitz_Kbar}
\end{align}
It is easy to check that such a construction leads to (\ref{eq:Pd_equal_QK}).
Its parameters are $\{\textbf{f}_d, \boldsymbol{\theta}_d, \boldsymbol{\Lambda}_d\}_d$, which can be trained through stochastic gradient descent (SGD) as usual. 

\textbf{Variant II. Relative (vanishing) attention with regular sampling} (\texttt{convSPE}). 
Due to their periodic structure, the covariance functions generated by Variant I are \emph{non-vanishing}. 
Yet, our framework is flexible enough to allow for vanishing covariance structures, which may be more desirable depending on the application \cite{wang2021on}. 

As opposed to Variant I, where we imposed a specific structure on $\mathcal{P}_d$, we will now follow an indirect approach, where $\mathcal{P}_d$ will be \emph{implicitly} defined based on our algorithmic construction. In this case, we assume that the signals are regularly sampled (typical in e.g.\ text, images, audio), and we will exploit the structure of Gaussian random matrices and basic properties of the convolution operation.

For ease of notation, we assume self attention, i.e.\ $\mathcal{M}=\mathcal{N}$. Let $\{\boldsymbol{\Phi_d^Q},\boldsymbol{\Phi}_d^K\}_d$ denote a collection of \emph{filters}, which will ultimately be learned from training data. The size and the dimension of these filters can be chosen according to the input data (i.e.\ can be vectors, matrices, tensors). We then propose the following procedure, which leads to a Toeplitz~$\textbf{P}_d$ by means of \textit{convolutions}:   
\begin{itemize}[leftmargin=*, noitemsep,topsep=0pt]
\item We first draw an $M\times R$ random matrix $\textbf{Z}_d$ with i.i.d.\ standard Gaussian entries. For multidimensional signals, $\textbf{Z}_d$ gathers $R$ random vectors, matrices, cubes, etc.
\item The desired $\overline{\textbf{Q}}_d$ and $\overline{\textbf{K}}_d$ are obtained by convolving $\textbf{Z}_d$ with respective filters $\boldsymbol{\Phi_d^Q}$ and $\boldsymbol{\Phi}_d^K$: 
\vspace{-0.1cm}\begin{align}
\overline{\textbf{Q}}_{d}=\textbf{Z}_{d}\ast\boldsymbol{\Phi}_{d}^{Q}\,,\text{ }\overline{\textbf{K}}_{d}=\textbf{Z}_{d}\ast\boldsymbol{\Phi}_{d}^{K}\,,\label{eq:SPE_conv}
\vspace{-0.3cm}
\end{align}

where $\ast$ denotes convolution with appropriate dimension (e.g.\ 1D, 2D or 3D). Using convolutions with finite filters ensures vanishing covariance, as proven in the appendix.
\end{itemize}

Due to the independence of the entries of $\textbf{Z}_{d}$, for large $R$, the product  $\textbf{Z}_{d} \textbf{Z}_{d}^\top /R$ will tend to the identity matrix. Given the fact the convolution operations in \eqref{eq:SPE_conv} can be equivalently expressed as a multiplication by triangular Toeplitz matrices constructed from the respective filters, it can be shown that, as $R \to \infty$, $\frac1{R}\overline{\textbf{Q}}_d \overline{\textbf{K}}_d^\top$ tends to the product of two triangular Toeplitz matrices. Hence, by using the properties of triangular Toeplitz matrices (cf.\ \citealt{kucerovskytoeplitz2016}, Theorem 2.4), we conclude that, as $R  \to \infty$, our construction yields a Toeplitz matrix $\textbf{P}_d$ as desired. 
\\This approach is parameterized by the filters$\{\boldsymbol{\Phi}_d^Q, \boldsymbol{\Phi}_d^K\}_d$, which will be learned from training data through SGD.

The variety of attention patterns $\mathcal{P}(m-n)$ that can be obtained directly depends on the kernel sizes, which is a classical result from signal processing~\cite{vetterli2014foundations}. Cascading several convolutions as in the VGGNet \cite{simonyan2014vggnet} may be a convenient way to augment the expressive power of this convolutional SPE variant. 

From a more general perspective, the two operations in~(\ref{eq:SPE_conv}) can be understood as producing PE through filtering white noise, which is the core idea we introduce for PE. Other classical signal processing techniques may be used like using \textit{infinite impulse response} filters. Such considerations are close to the ideas proposed in \cite{engel2020ddsp}.

To summarize, the core difference between the two proposed constructions (\ref{eq:toeplitz_Qbar}-\ref{eq:toeplitz_Kbar}) and (\ref{eq:SPE_conv}) lies in the behaviour of RPE beyond a maximum lag, implicitly defined through the frequencies $\textbf{f}_d$ for (\ref{eq:toeplitz_Qbar}-\ref{eq:toeplitz_Kbar}) and through the sizes of the filters for (\ref{eq:SPE_conv}). While the sinusoidal construction leads to a periodic RPE, the filtering construction leads to a vanishing RPE, which is called \textit{monotonic} in \cite{wang2021on}. Both may be the desired option depending on the application.

\subsection{Gated SPE}
\label{sec:gating}
Although RPE and the generalization (\ref{eq:PE_in_attention_domain_matrix_form}) we propose are  novel and efficient strategies to handle position information, it may be beneficial to also allow for attention coefficients that are computed without positional considerations, simply through $\left<\textbf{q}_m,\textbf{k}_n\right>$. As a general \textit{gating} mechanism, we propose to weight between positional and non-positional attention through a \textit{gate parameter} $\delta_d\in[0~1]$:
\begin{equation}
        \textbf{P}_d\equiv\mleft[\delta_d + \mleft(1-\delta_d\mright)\mathcal{P}_d\mleft(m,n\mright)\mright]_{m,n}\,.\label{eq:gating}
\end{equation}
This gating scheme can be implemented simply by augmenting $\overline{\textbf{Q}}_d$ and $\overline{\textbf{K}}_d$ generated as above through:
\vspace{-0.15cm}\begin{align}
    \overline{\textbf{q}}_{d,m}	\leftarrow\sqrt{1-\delta_{d}}\overline{\textbf{q}}_{d,m}+\sqrt{\delta_{d}}\boldsymbol{\epsilon}_{d}\,,\label{eq:gating_q}\\
\overline{\textbf{k}}_{d,m}	\leftarrow\sqrt{1-\delta_{d}}\overline{\textbf{k}}_{d,m}+\sqrt{\delta_{d}}\boldsymbol{\epsilon}_{d}\label{eq:gating_k}\,,
\vspace{-0.3cm}\end{align}
where $\boldsymbol{\epsilon}_d\in\mathbb{R}^R$ in (\ref{eq:gating_q}) and (\ref{eq:gating_k}) is the same and has i.i.d.\ standard Gaussian entries. 

In practice, we can share some SPE parameters across the network, notably across layers, to strongly reduce computing time and memory usage. In our implementation, \textit{sharing} means generating a single instance of $\overline{\textbf{Q}}$ and $\overline{\textbf{K}}$ for each head, on which a layer-wise gating is applied, before achieving PE through (\ref{eq:SPE_PE_Q}-\ref{eq:SPE_PE_K}). This is illustrated in Figure~\ref{fig:SPE_illustration}.

\begin{table*}
    \caption{Long-Range Arena results (higher scores are better). Mean and standard deviation of accuracy over three runs is reported, except for Performer with convolutional SPE, where only a single run was completed. %
    For comparison, the best result reported by \citet{tay2021long}, along with the name of the best-performing model (in parentheses), is included.}
    \vskip 0.1in  %
    \renewcommand{\arraystretch}{0.95}
    \newcommand{\nopm}[1]{\multicolumn{1}{r@{\hphantom{\,$\pm\,$}}}{{#1}}}%
    \centering%
    \scalebox{0.95}{
    \begin{tabular}{lr@{\,$\pm\,$}lr@{\,$\pm\,$}lr@{\,$\pm\,$}lr@{\,$\pm\,$}l}
    \toprule
     & \multicolumn{2}{l}{ListOps} & \multicolumn{2}{l}{Text} & \multicolumn{2}{l}{Retrieval} & \multicolumn{2}{l}{Image} \\
    \midrule
    Best result from \citet{tay2021long} & \nopm{37.27} & & \nopm{65.90}& & \nopm{59.59}& & \nopm{44.24}\\[-.5ex]  %
    \multicolumn{1}{l}{} & \multicolumn{2}{l}{\small (Reformer)}
     & \multicolumn{2}{l}{\small (Linear Trans.)} & \multicolumn{2}{l}{\small (Sparse Trans.)} & \multicolumn{2}{l}{\small (Sparse Trans.)}\\[.1ex]
    Linear Transformer-ReLU from \citeauthor{tay2021long} & \nopm{18.01} & & \nopm{65.40} & & \nopm{53.82} & & \nopm{42.77} \\ %
    \midrule
    Performer-softmax (\texttt{APE}) & {\tbf 17.80} &  0.00 &  62.58 &  0.22 &  59.84 &  1.46 &   41.81 &  1.16 \\
    Performer-softmax + \texttt{sineSPE}       &   17.43 &  0.32 &  62.60 &  0.50 &  60.00 &  1.20 &   41.12 &  1.70 \\
    Performer-softmax + \texttt{convSPE} &   \nopm{\tbf 17.80} &    &    \nopm{60.94} &    &  \nopm{57.22} &    &   \nopm{40.06} &    \\
    Linear Transformer-ReLU (\texttt{APE})          &   17.58 &  1.01 &  63.98 &  0.05 &  58.78 &  0.93 &   {\tbf 42.25} &  0.01 \\
    Linear Transformer-ReLU + \texttt{sineSPE} &   {\tbf 17.80} &  0.00 &  {\tbf 64.09} &  0.62 &  {\tbf 62.39} &  0.59 &   41.21 &  1.18 \\
    Linear Transformer-ReLU + \texttt{convSPE}  &    9.50 &  1.17 &  63.23 &  1.31 &  61.00 &  1.34 &   39.96 &  1.31  \\
    \bottomrule
    \end{tabular}
    }
    \label{tab:lra-results}
    \vskip -0.1in
\end{table*}

\section{Experiments}
\subsection{Long-Range Arena}
\textbf{Experimental setup.} We evaluate the proposed method in the Long-Range Arena (LRA; \citealp{tay2021long}), a benchmark for efficient Transformers, consisting of sequence classification tasks with a focus on long-range dependencies. We use the following tasks from this benchmark:
\vspace{-0.5em}\begin{itemize}[leftmargin=*, noitemsep,topsep=0pt]
    \item \emph{ListOps}: parsing and evaluation of hierarchical expressions. a longer variant of \citep{nangia-bowman-2018-listops};
    \item \emph{Text}: movie review sentiment analysis on the IMDB corpus \citep{maas-etal-2011-learning};
    \item \emph{Retrieval}: article similarity classification on the All About NLP (AAN) corpus \citep{Radev2013};
    \item \emph{Image}: object recognition on the CIFAR10 dataset \citep{Krizhevsky2009LearningML} represented as pixel sequences.
\end{itemize}
\vspace{-0.5em}
The tasks are challenging due to the large sequence lengths, deliberately increased by choosing a character-/pixel-level representation.
An overview of the tasks can be found in the appendix.
We do not include \emph{Pathfinder} (a synthetic image classification task) as we were unable to reproduce the results of \citeauthor{tay2021long}\ on this task, even through correspondence with the authors.

We evaluate SPE (the gated variant) on two efficient Transformer models: the (softmax) Performer \cite{choromanskiRethinkingAttentionPerformers2020a}, and a Linear Transformer \cite{katharopoulosTransformersAreRNNs} with a ReLU feature map, i.e.\ choosing $\phi(\cdot)=\max(0,\cdot)$ element-wise in (\ref{eq:features_linear_attention}).\footnote{A model named `Performer' is reported by \citeauthor{tay2021long}, but communication with the authors revealed it to be in fact equivalent to our Linear Transformer-ReLU, as it does not use random features.
To avoid confusion, we refer to this model as such herein.}
It should be noted that the ReLU feature map does not approximate the softmax kernel, which SPE is designed for (see assumption~\ref{eq:PE_in_attention_domain}).
Nevertheless, it is possible to use SPE with any feature map in practice, allowing us to include Linear Transformer-ReLU as an interesting test of generalization to alternative kernels.

We adopt the configuration of \citeauthor{tay2021long}, only changing the PE and the batch sizes/learning rates to allow training on limited hardware with similar results. All other hyperparameters are kept identical to the original LRA.
It is worth noting that the \emph{Image} models are different from the rest in that they employ a single-layer network and only use the first position for prediction, dramatically limiting their ability to benefit from relative positional information.

Since we observe some variation between different runs, we train and evaluate each model 3 times (except for Performer with convolutional SPE, which is computationally more costly) and report the mean and standard deviation of the results.

The results of the benchmark are given in Table \ref{tab:lra-results}.
The accuracies achieved by the baseline Linear Transformer-ReLU (\texttt{APE}) are similar to or surpass those reported by \citeauthor{tay2021long}, which is a clear validation of our experimental setup. 

\textbf{Discussion.} Results on ListOps are poor overall, with accuracies around $17\,\%$. This complies with \citet{tay2021long}, who reasoned that ``kernel-based models [e.g.\ Performer, Linear Transformers] are possibly not as effective on hierarchically structured
data,'' leaving room for improvement. We also hypothesize this is largely due to some known issues with the training data for this task, which unfortunately have not been fixed at the time of this writing.\footnote{Currently, the official data loader for ListOps inadvertently strips some characters from the input sequences.}

Regarding performance of SPE, we first notice that the \texttt{sineSPE} variant yields the best results on three tasks, which is a strong achievement and validates our approach, especially considering the difficulty of this evaluation benchmark. While it is only marginally better than APE for \emph{ListOps} and \emph{Text}, it is worth mentioning that \texttt{sineSPE} combined with the Linear Transformer-ReLU yields an accuracy improvement of  $\sim${}$3\,\%$ on \emph{Retrieval} compared to the best result obtained by \citet{tay2021long}. %

Regarding \texttt{convSPE}, its performance in the LRA is not as remarkable as it is for the music generation experiment reported later in section~\ref{sec:unconditional-pop}. This mitigated result appears somewhat in contradiction with the discussion found in \citet{wang2021on}, which presents vanishing attention as a desirable property of PE. On the contrary, we empirically observe that our non-vanishing sinusoidal version \texttt{sineSPE} does behave better in these particular tasks.

Finally, the superior results of \texttt{APE} on \emph{Image} are not unexpected, given the limited ability of these models to exploit relative positions. On the contrary, the relatively good performance of SPE on this task is in fact remarkable, especially considering that the baseline systems for this task use \emph{learnable} APE. %

As we will see later in our music generation experiments, there are tasks where our proposed SPE clearly yields remarkable improvements. Here in the LRA, we notice that it does not result in an obvious and systematic boost in performance. This raises interesting considerations:

(i)~\emph{The variance of the Monte Carlo estimator might be problematic.} We are enthusiastic about the elegant formulation of stochastic feature maps as in the Performer, which was a strong inspiration. Still, we must acknowledge that their computation relies on a Monte Carlo estimator~\eqref{eq:E_as_a_product}. We suspect that the variance of the estimator might play a role in the final performance in large dimensions, which opens up the direction of exploring variance-reduced estimation methods, rather than plain Monte Carlo.

(ii)~\emph{LRA tasks might not benefit from strong (R)PE schemes.} The LRA was designed to compare Transformer \textit{architectures}, filling a gap in this domain and standing as the \textit{de facto} standard, justifying our choice. Still, although PE is known to be important in many cases, it is not known whether it is so in the LRA tasks. %
We feel that there is room for such a specialized comparison, which is scheduled in our future work, possibly leading to new long-range tasks where PE is critical.

\subsection{Pop Piano Music Generation}
\label{sec:unconditional-pop}

In our music generation experiments (this subsection and section \ref{sec:music-continuation}), music is represented as sequences of symbols (tokens) and a Performer \cite{choromanskiRethinkingAttentionPerformers2020a} is used as an autoregressive language model, which predicts a probability distribution over the next token given the past context. At test time, a new sequence is generated by iteratively sampling the next token, as commonly done in text generation.

\textbf{Experimental setup}. We train Performers for music generation, with 24 layers and 8 heads per layer on a dataset composed of $1\,747$ pop piano tracks, encoded using the recently proposed \emph{Revamped MIDI-derived format} (REMI; \citealp{huang2020pop}).
The sequences are composed of \textit{metrical} tokens: \texttt{bar}, \texttt{subbeat}, and \texttt{tempo}, which represent musical timing; and \textit{note} tokens: \texttt{chord}, \texttt{pitch}, \texttt{duration}, and \texttt{volume}, which describe the musical content (see the appendix for more details). We hold out 5\% of the songs as the validation set.

We train the models with sequence length $N=2\,048$, corresponding to $\sim${}$1$ minute of music. The only difference between our models is the PE strategy. We consider baseline \texttt{APE}, as well as SPE: sinusoidal or convolutional, with or without gating, resulting in 5 different models.

\textbf{Results and discussion}. For qualitative assessment, we first display in Figure~\ref{fig:attention_illustration} one attention pattern for each PE model: \texttt{APE} and (gated) \texttt{sineSPE}/\texttt{convSPE}, obtained as an average over $20$ from-scratch generations for a chosen (layer, head). More similar plots can be found in appendix. Interestingly, we notice that for early layers, \texttt{APE} attention does not go much beyond training sequence length. This behaviour is not found in SPE variants, which consistently attend to all positions. Another remarkable feature of the proposed model (only displayed in the appendix) is that \textit{gating} as described in section~\ref{sec:gating} visually disables PE altogether for some layers/heads, in which case attention is global.

\begin{figure}
    \centering
    \includegraphics[width=0.95\linewidth]{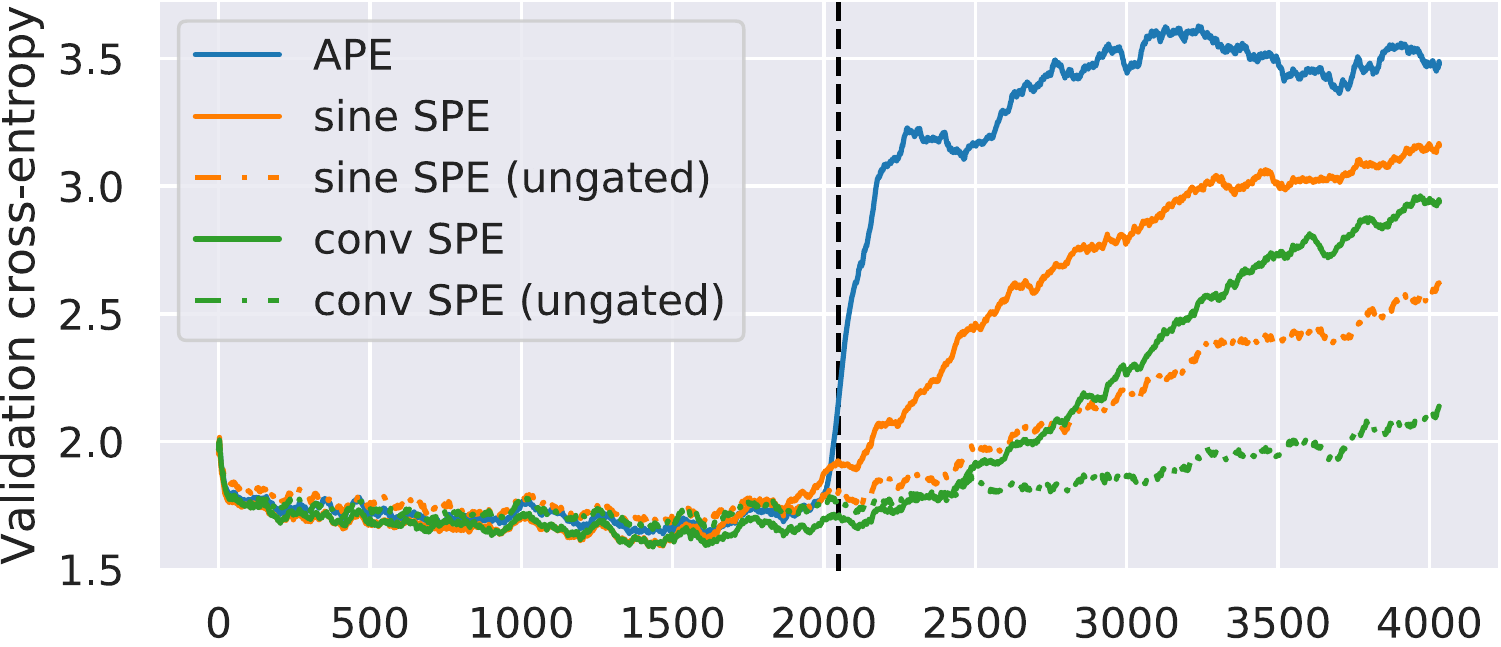}
    \vspace{-0.4cm}
\caption{Validation cross-entropy vs. token position on pop piano music generation task. (lower is better; the \textbf{black} vertical line indicates the maximum position to which the models are trained.)\label{fig:music-pop-crossentropy}}
\vspace{-0.1cm}
\end{figure}

\begin{figure}
    \centering
    \includegraphics[width=0.95\linewidth]{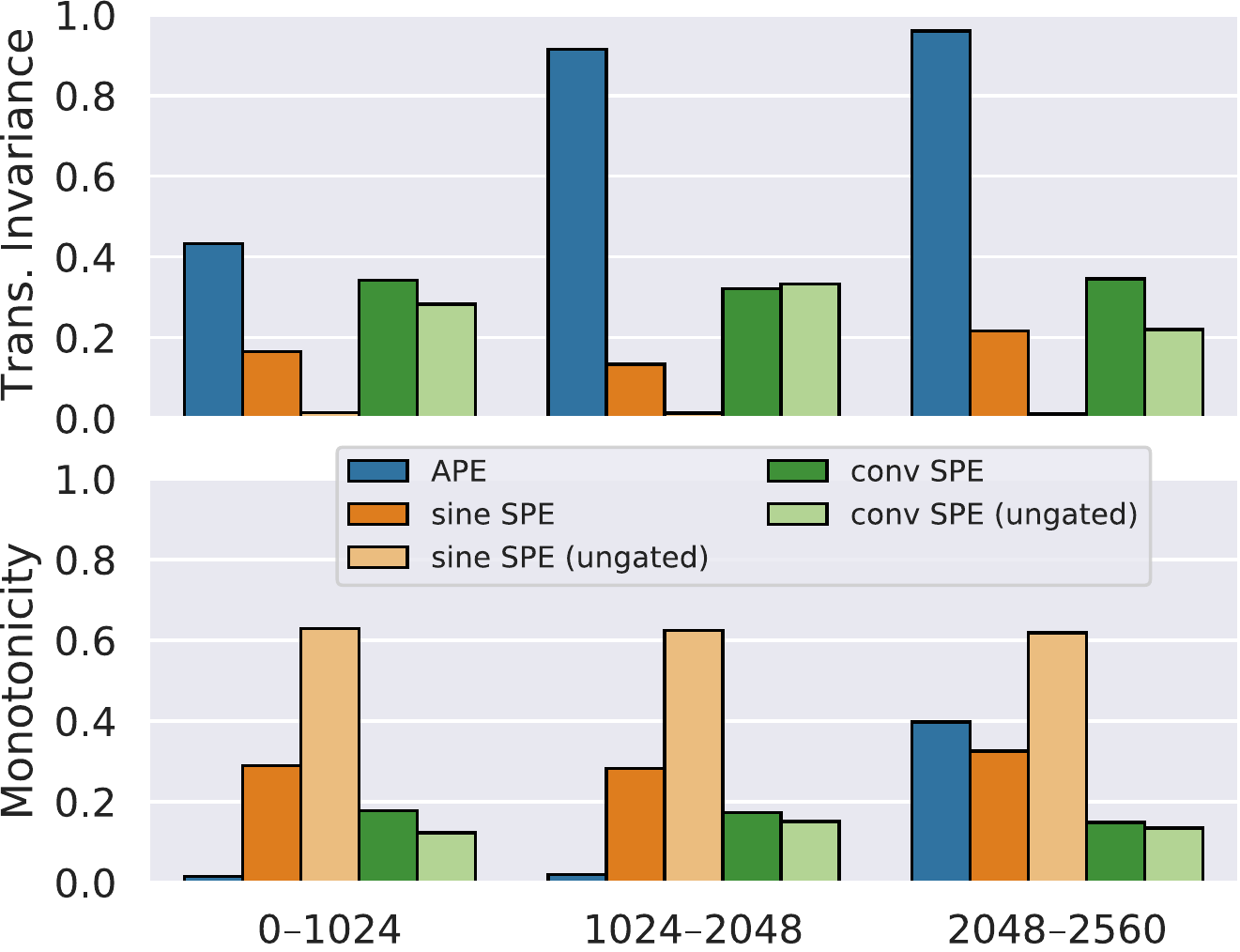}
    \vspace{-0.4cm}
\caption{PE evaluation metrics \cite{wang2021on} for the pop piano music generation task in the $1$st layer (lower is better), w.r.t.\ query positions. Training sequence length is $2\,048$. Only query-key offsets ${<}128$ are considered here. See appendix for details.\label{fig:music-pop-PEmetrics}}
\vspace{-0.1cm}
\end{figure}

Since the literature suggests that RPE improves generalization performance~\cite{shawSelfAttentionRelativePosition2018,zhou2019improving, rosendahlAnalysisPositionalEncodings}, we display validation cross-entropy computed with teacher forcing \cite{williams1989learning} in Figure~\ref{fig:music-pop-crossentropy}, as a function of the 
target token position.
The values would indicate how well the models predict the token at a certain position given the preceding tokens, for tracks in the validation set.
We notice that all SPE variants, especially \texttt{convSPE}, behave much better than \texttt{APE} for token positions beyond $2\,048$. This suggests that SPE inherits this celebrated advantage of RPE \citep{huangMusicTransformer2018} while being applicable to much longer sequences.%

Recently, \citet{wang2021on} defined metrics for the evaluation of PE, suggesting that \textit{translation invariance} and \textit{monotonicity} are desirable properties. 
The former states that the distances of two arbitrary $\tau$-offset position embeddings should be identical, while the latter states that neighboring positions should be assigned with position embeddings that are closer than faraway ones.
Following their \textit{identical word probing} methodology, we report these metrics in Figure~\ref{fig:music-pop-PEmetrics}. As expected, SPE variants greatly outperform \texttt{APE} in terms of \textit{translation invariance}. However, \textit{monotonicity} does not seem a very relevant criterion in our music application, as can be seen when comparing scores in Figures~\ref{fig:music-pop-crossentropy} and \ref{fig:music-pop-PEmetrics}. It seems that music modeling can benefit from non-vanishing attention patterns. In any case, SPE scores are remarkably stable across positions, contrarily to \texttt{APE}, which rapidly degrades beyond the training length.

\subsection{Groove Continuation}
\label{sec:music-continuation}

\begin{figure}[t]
\centering
\includegraphics[width=0.94\linewidth]{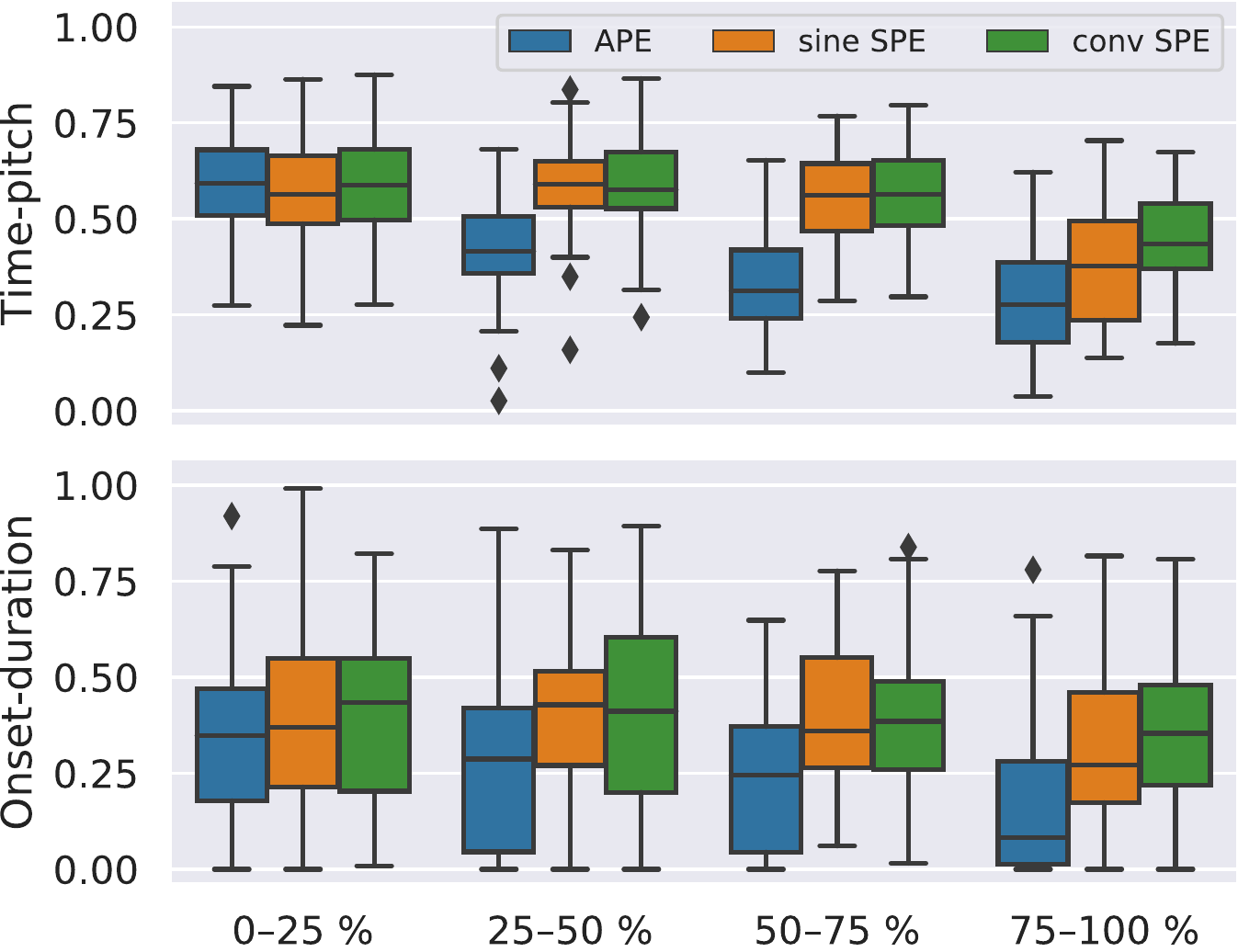}
\vspace{-0.3cm}
\caption{Musical style similarity for groove continuation (higher is better) between output and initial prompt through two musically-motivated metrics, as a function of time in the output. Each data point corresponds to a single musical style.\label{fig:music-continuation}}
\end{figure}

In this experiment, 
we %
evaluate Performers
on a \emph{groove continuation} task.
After training on a dataset where each example has a uniform style (`groove'), we prime the model with a short \emph{prompt} (2-bar musical fragment) and let it generate a continuation. We then observe whether the generated continuation matches the style of the prompt.

\textbf{Experimental setup.} The models (24-layer Performers with 8 attention heads) are trained on an accompaniment dataset comprising $5\,522$ samples in $2\,761$ different musical styles, encoded in a token-based format adopted from \citet{cifka:hal-02923548} and detailed in the appendix.
All SPE-based models use gating in this experiment.
Unlike the previous experiment, which leverages long training sequences, we consider training sequences of length $N=512$, corresponding to 2--10 bars. At test time, the model is prompted with 2 bars in a style not seen during training and new tokens are sampled to complete the sequence to a length of $1\,024$, %
i.e.\ twice the training length.

We use two musically motivated \emph{style similarity} metrics~-- \emph{time-pitch} and \emph{onset-duration} proposed by \citet{cifka-ismir2019,cifka:hal-02923548}~-- to quantify the similarity of the generated continuation to the prompt.
When listening to the generated music, we perceptually notice a drift in quality along time. %
For this reason, we divide each generated sample into four successive chunks of identical duration and evaluate them independently.
The results are displayed in Figure~\ref{fig:music-continuation}.

\textbf{Discussion.} We clearly see that SPE substantially outperforms \texttt{APE} in both metrics.
Although \texttt{APE} visibly does manage to generate close to the desired style at the beginning of the sequence, this similarity strongly degrades over time. Both  \texttt{sineSPE} and \texttt{convSPE} are much more stable in this regard, confirming the result from section \ref{sec:unconditional-pop} that SPE extrapolates better beyond the training sequence length. %
This matches our informal perceptual evaluation.\footnote{Examples: \url{https://cifkao.github.io/spe/}}

This experiment suggests that exploiting a local neighborhood is a robust way to process long sequences. This could appear as contradicting the use of long-range Transformers, but we highlight that \textit{gating} is used here, enabling some heads to exploit long term-attention independently from position. Further comparisons with local attention schemes (e.g.\ \citealp{daiTransformerXLAttentiveLanguage2019,hofstatter2020local}) could be interesting, although they were not included here due to~\citet{tay2021long} suggesting that they are clearly inferior, at least in the LRA setting.

\section{Related Work}
\label{sec:related_work}

This paper is concerned with PE \cite{sukhbaatarEndToEndMemoryNetworks2015}, as a way to embed the position of each token as part of its features. This idea is a core ingredient for many subsequent groundbreaking studies \cite{gehringConvolutionalSequenceSequence2017,vaswaniAttentionAllYou}, and has been the actual topic of many investigations.

\vspace{-1pt}
\textbf{Absolute Positional Encoding (APE)} based on sinusoids from \citet{vaswaniAttentionAllYou} is the most widely used for Transformer-like architectures. However, PE $\overline{\textbf{q}}(n)$ and $\overline{\textbf{k}}(n)$ in (\ref{eq:additive_PE_keys_domain}) can also be trained as in BERT \cite{devlinBERTPretrainingDeep2019, liuRoBERTaRobustlyOptimized2019}. Although the original Transformer only includes PE at the input layer, it may be included at all layers \cite{dehghaniUniversalTransformers2019,lanALBERTLiteBERT2020}.

\vspace{-1pt}
\textbf{Relative positional encoding (RPE;}  \citealp{shawSelfAttentionRelativePosition2018}\textbf{)} is a way to leverage \textit{relative positions}. It came with a $\mathcal{O}(N^2 D)$ space complexity, which was reduced to $\mathcal{O}(N^2)$ in \citet{huangMusicTransformer2018, heDeBERTaDecodingenhancedBERT2020}. Considering log-distances was proposed in \citet{raffelExploringLimitsTransfer2020}. Several variants for RPE were introduced \cite{huang2020improve, wang2021on}. They all apply learned RPE in the attention domain.
Using fixed embedding functions was also considered for RPE \cite{phamRelativePositionalEncoding2020}, and masking RPE is used in \citet{kimTGSATransformerGaussianweighted2020} to promote local attention.

\vspace{-1pt}
\textbf{Keys-domain vs attention domain.} Doing PE in the keys domain introduces position-content cross terms that are advocated as noisy and not beneficial in \citet{keRethinkingPositionalEncoding2020} and replaced by \textit{Untied} attention, i.e.\ PE in the attention domain. This is also called \textit{disantangled attention} in \citet{heDeBERTaDecodingenhancedBERT2020} and already proposed in \citet{tsai2019transformer} through \textit{separable} content-position attention kernels. All of these studies require the explicit computation and storage of $\textbf{A}$.

\vspace{-1pt}
\textbf{Non-integer positions} were considered for structured inputs. Tree-based PE was proposed both for APE \cite{shivNovelPositionalEncodings,xiaoLatticeBasedTransformerEncoder2019, maImprovingNeuralMachine2019} and RPE \cite{ehimeuniversityjapanDependencyBasedRelativePositional2019}. Positional encoding of robots within arbitrary polygons is found in \citet{bosePositionalEncodingRobots2019}.

\vspace{-1pt}
\textbf{Dynamical models for PE}. Attention for machine translation was introduced in \citet{bahdanauNeuralMachineTranslation2016}, which was retrospectively understood in \citet{keRethinkingPositionalEncoding2020} as using recurrent neural nets (RNN) for PE. In~\citet{chenBestBothWorlds2018}, the hidden states of encoder RNNs are said to contain enough position information to skip explicit PE. \citet{neishiRelationPositionInformation2019} builds on this view, but explicitly describes the idea for the first time. Their contribution is to replace the additive PE in~(\ref{eq:additive_PE_keys_domain}) by an RNN. In the same vein, \citet{liuLearningEncodePosition2020} generates PE using (neural) ordinary differential equations.

\vspace{-1pt}
\textbf{Convolutional contexts.}  Our \texttt{convSPE} variant involves convolving random noise. First, this can be related to~\citet{mohamed2019transformers}, who use convolutional neural networks for queries and keys computation. Second, the connections between convolutions and stationary processes have recently been highlighted by \citet{xu2020positional} as enforcing PE.

\vspace{-1pt}
\textbf{Multiplicative PE.} Various levels of content-position interactions are formalized in \cite{tsai2019transformer}. Multiplicative strategies were proposed for both RPE \cite{huang2020improve} and APE \cite{daiTransformerXLAttentiveLanguage2019}. The latter was generalized in \citet{tsai2019transformer}. All these require the explicit computation of the attention matrix. \citet{wangEncodingWordOrder2020} presents a scheme that is close to our sinusoidal variant, but without the stochastic part that is the key to go from (\ref{eq:E_as_a_sum_of_product}) to (\ref{eq:E_as_a_product}).

\vspace{-1pt}
\textbf{The limits of APE and RPE} were highlighted by some authors. In \citet{wangWhatPositionEmbeddings2020}, the best performing models exploit both absolute \textit{and} relative positions. In \citet{irieLanguageModelingDeep2019} and \citet{tsai2019transformer}, it is found that removing APE altogether in the causal decoder part of Transformer-based architectures leads to comparable/better performance. It is also not clear which one is best between incorporating PE in the raw input signal (and hence propagating it through the \textit{value} entries) or using it anew on the queries and keys only, as we do. Our choice is backed by \citet{tsai2019transformer}.

\section{Conclusion}

We propose a new Stochastic Positional Encoding (SPE), based on filtering random noise. As we show, the procedure generalizes relative PE and is a principled means to enforce any prescribed (but trained) cross-covariance structure, which we demonstrated should be the central concern in dot-product attention.
In our experiments, we show that SPE brings an interesting gain in performance for large-scale transformer models \cite{choromanskiRethinkingAttentionPerformers2020a, katharopoulosTransformersAreRNNs}, as compared to classical (sinusoidal) PE. This was expected, because  RPE \cite{shawSelfAttentionRelativePosition2018} is often advocated as beneficial. However, no way to incorporate it for long sequences was available so far and this is the core contribution of this paper. 
The natural future directions for our study are (i) \emph{Signal-dependent PE} that incorporates the input sequence as an additional input for SPE, (ii) \emph{Nonstationary PE} that utilizes both relative and absolute positions, (iii) Extending our approach to \emph{arbitrary attention kernels}, e.g.\ defined implicitly through their (random) mappings as in   (\ref{eq:FAVOR}). Indeed, SPE as it is presented here holds theoretically for dot-product attention kernels only, but our results given in Table~\ref{tab:lra-results} suggest that this generalizes, asking an interesting research question.

\section*{Acknowledgements}
This work was supported by the European Union’s Horizon 2020 research and innovation programme under the Marie Skłodowska-Curie grant agreement No.\ 765068 (MIP-Frontiers) and in part by the French government under management of Agence Nationale de la Recherche as part of the ``Investissements d’avenir'' program, reference ANR-19-P3IA-0001 (PRAIRIE 3IA Institute).

We would like to thank Yi Tay, Mostafa Dehghani and Philip Pham for their help with troubleshooting the Long-Range Arena, and Krzysztof Choromanski for clarifications about the Performer.

\bibliography{transformers_and_positional_encoding,various_linear_algebra_things,gaussian_simulations,music,misc_ml}
\bibliographystyle{icml2021}

\clearpage
\icmltitlerunning{Supplementary Material: Relative Positional Encoding for Transformers with Linear Complexity}
\twocolumn [
\icmltitle{Relative Positional Encoding for Transformers with Linear Complexity\\[1ex]Supplementary Material}

]

\appendix
\section*{Introduction}
This document comprises additional information that could not be included in the paper due to space constraints. It is structured as follows. In appendix \ref{sec:theory}, we provide some further theoretical developments. In appendix~\ref{sec:LRA}, we detail the experimental setup on the Long Range Arena. In appendix~\ref{sec:music_gen}, we detail our music generation experiments. Finally, we provide additional results in appendix~\ref{sec:additional}.

Our source code is available at:
\begin{center}
\url{https://github.com/aliutkus/spe/}
\end{center}
See also the companion website:
\begin{center}
\url{https://cifkao.github.io/spe/}
\end{center}

\section{Theory}
\label{sec:theory}
\subsection{Convolutional SPE Leads to Vanishing Attention}

\begin{figure}[h]
     \centering
    \noindent\fbox{\begin{minipage}[t]{0.6\columnwidth}%
    \includegraphics[width=0.8\columnwidth]{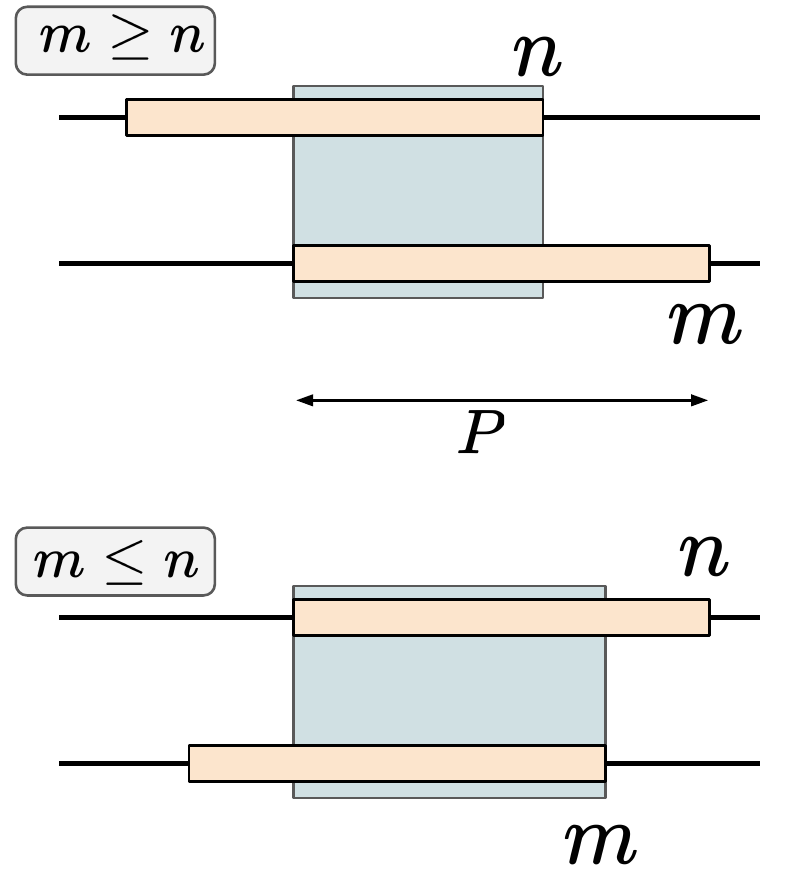}
    \end{minipage}}
\caption{If $\boldsymbol{\Phi}_d^Q$ and $\boldsymbol{\Phi}_d^K$ have length $P$, $\overline{Q}_d$ and $\overline{K}_d$ for convolutional SPE depend on the noise $Z_d$ over the intervals $[m-P:m]$ and $[n-P:n]$, respectively. Their correlation depends only on the shaded area, due to whiteness of $Z_d$. Whenever $|m-n|>P$, the two signals are uncorrelated. \label{fig:MA_illustration}}
\end{figure}

In the main document, we claim that the convolutional variant leads to vanishing attention. We shortly prove this claim here. For ease of notation, the proof is given in the 1D case, but extends trivially to higher dimensions. The core idea is illustrated in Figure~\ref{fig:MA_illustration}. Convolutional SPE yields:
\vspace{-0.3cm}
\begin{align*}
\overline{Q}_{d}\mleft(m,r\mright) & =\sum_{p=0}^{P}Z_{d}\mleft(m-p,r\mright)\phi_{d}^{Q}\mleft(p\mright),\\
\overline{K}_{d}\mleft(n,r\mright) & =\sum_{p=0}^{P}Z_{d}\mleft(n-p,r\mright)\phi_{d}^{K}\mleft(p\mright),
\vspace{-0.3cm}
\end{align*}
where $Z_d$ is a white Gaussian noise process, i.e. $\mathbb{E}[Z_d(m,r)Z_d(m',r)]=\delta_{mm'}$.
Omitting the dependency on $r$ for notational convenience (all realizations are independent), we can compute the positional attention as:

\vspace{-0.3cm}\begin{align*}
\mathcal{P}_{d}\mleft(m,n\mright)
&=\mathbb{E}\left[\overline{Q}_{d}\mleft(m\mright)\overline{K}_{d}\mleft(n\mright)\right]\\
&=\mathbb{E}\left[\sum_{p,\tau=0}^{P}z_{d}\mleft(m-\tau\mright)z_{d}\mleft(n-p\mright)\phi_{d}^{Q}\mleft(\tau\mright)\phi_{d}^{K}\mleft(p\mright)\right]\\
&=\mathbb{E}\left[\sum_{p=0}^{P}z_{d}\mleft(n-p\mright)^{2}\phi_{d}^{Q}\mleft(p+m-n\mright)\phi_{d}^{K}\mleft(p\mright)\right]\\
&=\sum_{p=0}^{P}\phi_{d}^{Q}\mleft(p+m-n\mright)\phi_{d}^{K}\mleft(p\mright),
\vspace{-0.4cm}\end{align*}
where only the $(p,\tau)$ values such that $n-p=m-\tau$ remain, all other cross terms $\mathbb{E}[Z_d(m)Z_d(m'\neq m)]$ disappearing due to whiteness of $Z_d$. Filters are taken as $0$-valued outside of $[0:P]$. As can be seen, whenever $|m-n|>P$, we get $\mathcal{P}_d(m,n)=0$, because $\phi_d^K(p+(m-n))=0$.\qed

\subsection{Complexity}
In this section, we detail the additional complexity caused by the proposed SPE method.
\vspace{-0.1cm}\begin{itemize}[leftmargin=*, noitemsep,topsep=0pt]
    \item \textbf{Sinusoidal SPE} first requires the computation of the modulation matrices $\boldsymbol{\Omega}$ for each feature dimension $d=1\dots D$, which has a $\mathcal{O}(2NK)$ complexity. Then, this matrix must be multiplied by the noise matrix $\textbf{Z}_d$ with shape $2K\times R$, leading to an overall complexity of $\boldsymbol{\mathcal{O}(DRNK^2)}$. Since $K$ is typically very small in our experiments, \texttt{SineSPE} can be seen as quite light in terms of both time and space complexity. 
    \item \textbf{Convolutional SPE} involves drawing a new noise signal $\textbf{z}_{d,:,r}$ of length $N$ for each $d$ and $r$, and convolving it with the filters $\boldsymbol{\phi}_d^Q$ and $\boldsymbol{\phi}_d^K$, whose length is written $P$.
    \\In the 1D case, this leads to an overall time complexity of $\boldsymbol{\mathcal{O}(DRNP)}$, which can be replaced by $\mathcal{O}(DRN\log N)$ when operating the convolutions in the frequency domain, which is advantageous for long filters.
    \\In higher dimensions, say 2D, this becomes $\mathcal{O}(DRN_1N_2P_1P_2)$ in the original domain and $\mathcal{O}(DRN_1N_2\log N_1\log N_2)$ in the frequency domain, where $(N_1, N_2)$ and $(P_1, P_2)$ are the shapes of noise and filters, respectively.
    \item The bottleneck of \textbf{gating} %
    is the generation of random noise $\boldsymbol{\epsilon}_d$, which has complexity $\mathcal{O}(DR)$.
\end{itemize}

Note that this complexities of course must be multiplied by the number of heads considered, up to 8 in our experiments.

As can be seen, the complexities of the sinusoidal and convolutional variants are similar, depending on the length $P$ of the filters and the number $K$ of sinusoids.
\\Still, other aspects come into the play. First, the convolutional version requires generating noise whose size scales as $N$, while the sinusoidal version requires much smaller $2K$-large noise matrices. Second, only a very small number of sinusoids was required in our experiments, whereas the convolutional version required longer contexts, so that we often had $2K\ll P$ in practice. Finally, although this may change in the near future, deep learning frameworks like PyTorch do not easily integrate convolutions in the frequency domain.

\textbf{Sample-wise noise sharing.} In practice, SPEs do not need to be drawn anew for each example.
The most straightforward trick to reduce memory and computational footprint of the method is to share $\overline{\textbf{Q}}$ and $\overline{\textbf{K}}$ among all examples in each mini-batch, as we do in all our experiments.
This can bring significant memory savings when SPE is used as a drop-in addition to networks trained with large batch sizes.

\section{Experimental Setup: Long-Range Arena}
\label{sec:LRA}
\begin{table*}
    \caption{Long-Range Arena classification tasks used in this paper.}
    \vskip 0.1in  %
    \centering
    \begin{tabular}{llllrr}
    \toprule
    Name & Dataset & Input & Length & Goal & \# classes \\
    \midrule
    ListOps & ListOps & expression with operations on lists of digits & 2\,k & evaluate expression & 10 \\
    Text & IMDB & movie review as byte string & 8\,k & classify sentiment & 2 \\
    Retrieval & AAN & pair of articles as byte strings & $2\times4\text{\,k}$  & detect citation link& 2 \\
    Image & CIFAR10 & 8-bit gray-scale $32\times32$ image as byte string & 1\,k & recognize object & 10 \\
    \bottomrule
    \end{tabular}
    \label{tab:lra-tasks}
\end{table*}

An overview of the Long-Range Arena \cite{tay2021long} tasks is given in \cref{tab:lra-tasks}. We do not include \emph{Pathfinder} (a synthetic image classification task) or its harder variant \emph{Pathfinder-X} in this paper as we were unable to reproduce the results of \citeauthor{tay2021long}\ on this task.
All the datasets are described in detail in \citeauthor{tay2021long}\ and available from the official LRA repository.\footnote{\url{https://github.com/google-research/long-range-arena}}

In all LRA experiments, we employ gated SPE with $R\in\{32,64\}$. %
We consistently use $K=10$ for sinusoidal (periodic) SPE and filters of length 128 for convolutional SPE. For convolutional SPE, we share $\overline{\textbf{Q}}$ and $\overline{\textbf{K}}$ across all layers (but not across attention heads); for sinusoidal SPE, $\overline{\textbf{Q}}$ and $\overline{\textbf{K}}$ are unique to each layer and head; in both cases, layer-specific gating is employed.
Baseline experiments employ the same absolute positional encodings as \citeauthor{tay2021long} (learnable APE for Image and sinusoidal APE for the remaining tasks). In models employing SPE, APE is removed.

The numbers of parameters of the models presented in the main document are shown in Table \ref{tab:lra_num_params}.
We can see that SPE-based models have at most 3.1\,\% more parameters than the baselines.
In the Image column, the numbers for SPE-based models are about 50\,\% lower due to the fact that the baselines on this task employ learnable APE.

\begin{table*}
\caption{Numbers of parameters of LRA models, identical for both Performer-softmax and Linear Transformer-ReLU.}
\vskip 0.1in  %
\centering
\begin{tabular}{lrrrr}
\toprule
& ListOps &  Text &  Retrieval &  Image \\
\midrule
Baseline (APE) & 19\,982\,858 & 3\,486\,722 & 1\,087\,618 &  248\,458 \\
+ \texttt{sineSPE} & 20\,078\,090 & 3\,518\,466 & 1\,103\,490 &  119\,242 \\
+ \texttt{convSPE} & 20\,117\,002 & 3\,553\,282 & 1\,120\,898 &  133\,706 \\
\bottomrule
\end{tabular}
\label{tab:lra_num_params}
\end{table*}

We use code from the official LRA repository, including the authors' Transformer implementation, modified as necessary to incorporate SPE.
We keep the same training configuration as provided by the LRA authors, but decrease the batch sizes (from 256 to 96 for Image and from 32 to 8 for the rest) and learning rates so as to fit within 16~GB of GPU memory.
Our modified code and configuration files are available in our source code repository.

\subsection{Resource usage}
The typical training times of the LRA models are displayed in Table \ref{tab:lra_training_time}.
Note that the times may not be comparable across models or tasks due to evaluation (which may be time-consuming) being done more frequently in some runs than others.

The total training time was 1\,405\,h (189 runs in total), out of which 273\,h (61 runs) were spent on attempts to reproduce the results of \citet{tay2021long} using Performer-softmax, Linear Transformer-ReLU and vanilla Transformer.
Some of these preliminary experiments were distributed over 1--3 Tesla V100 GPUs with 32\,GB of memory each.
The final models were all trained on a single Tesla V100 or P100 GPU with 16\,GB of memory.

\begin{table*}
\caption{Training times for LRA models (hours). Numbers in parentheses are from Tesla P100 GPUs, the rest from Tesla V100 GPUs.}
\vskip 0.1in  %
\centering
\begin{tabular}{lrrrr}
\toprule
& ListOps &  Text &  Retrieval &  Image \\
\midrule
Performer-softmax &      1.1 &  4.8 &  1.2 &      4.8 \\
Performer-softmax + \texttt{sineSPE} &      \llap{(}4.2\rlap{)} & 11.7 &  2.9 &      5.0 \\
Performer-softmax + \texttt{convSPE} &      8.9 & 23.2 & 21.9 &      5.3 \\
Linear Transformer-ReLU &      0.6 &  \llap{(}3.2\rlap{)} &  0.7 &      4.8 \\
Linear Transformer-ReLU + \texttt{sineSPE} &      2.0 &  6.8 &  2.1 &      5.0 \\
Linear Transformer-ReLU + \texttt{convSPE} &     15.0 & 18.6 & 19.0 &      5.3 \\
\bottomrule
\end{tabular}
\label{tab:lra_training_time}
\end{table*}

\section{Experimental Setup: Music Generation}
\label{sec:music_gen}
Our music Performers are implemented using the \texttt{pytorch-fast-transformers} package,\footnote{\url{https://github.com/idiap/fast-transformers}}
modified as necessary to incorporate SPE.
The modified code and configuration files are available in our code repository.

All models have 24 layers with model dimension 512, 8 attention heads and 2\,048 feed-forward units, which amount to $\sim$80 million trainable parameters.
In models that use SPE, $\overline{\textbf{Q}}$ and $\overline{\textbf{K}}$ are shared across all layers (but not across attention heads); layer-specific gating is employed for models trained with gated SPE.

The models are trained with the Adam optimizer. We schedule the learning rate with linear warmup, followed by cosine decay. Full details of hyperparameters can be found in the provided configuration files.

\subsection{Pop Piano Music Generation}
\label{sec:pop-generation}

\begin{table}[h]
    \caption{Resource usage of models trained on pop piano music generation, on a Tesla V100 GPU with 32GB of memory. \# of epochs and time to the checkpoint with the lowest validation loss are displayed. (\texttt{ug}: trained without SPE gating.)}
    \vskip 0.1in  %
    \centering
    \begin{tabularx}{\linewidth}{lRRR}
    \toprule
         & \multicolumn{1}{c}{\# epochs} & \multicolumn{1}{c}{Time} & \multicolumn{1}{c}{Memory}\\
    \midrule
    \texttt{APE} & 72 & 9.74\,h & 14.34\,GB \\
    \texttt{sineSPE} & 78 & 17.92\,h & 29.80\,GB \\
    \texttt{sineSPE} (\texttt{ug}) & 78 & 16.31\,h & 18.29\,GB \\
    \texttt{convSPE} & 80 & 28.02\,h & 30.01\,GB \\
    \texttt{convSPE} (\texttt{ug}) & 68 & 24.76\,h & 18.99\,GB \\
    \bottomrule
    \end{tabularx}
    \label{tab:pop-resource}
\end{table}

\begin{table}
    \caption{Validation cross-entropy for models trained for pop piano music generation (mean and standard deviation) over all sequences. (\texttt{ug}: trained without SPE gating). \textit{Trained}: pos $\leq$ 2\,048, \textit{Extrapolation}: 2\,048 $<$ pos $\leq$ 3\,072.}
    \vskip 0.1in  %
    \centering
    \begin{tabularx}{\linewidth}{lCC}
    \toprule
      \multicolumn{1}{r}{\textit{Positions}} & \textit{Trained} & \textit{Extrapolation} \\
    \midrule
    \texttt{APE} & 1.721 $\,\pm$ 0.148 & 3.215 $\,\pm$ 0.200 \\
    \texttt{sineSPE} & 1.694 $\,\pm$ 0.148 & 2.396 $\,\pm$ 0.359 \\
    \texttt{sineSPE} (\texttt{ug}) & 1.754 $\,\pm$ 0.146 & 1.965 $\,\pm$ 0.170 \\
    \texttt{convSPE} & {\tbf1.685} $\,\pm$ 0.151 & 1.932 $\,\pm$ 0.225 \\
    \texttt{convSPE} (\texttt{ug}) & 1.733 $\,\pm$ 0.145 & {\tbf1.805} $\,\pm$ 0.163 \\
    \bottomrule
    \end{tabularx}
    \label{tab:remi-gen-perf}
\end{table}

\paragraph*{Training data.}
The pop piano MIDI dataset we use is 
derived from %
the one provided in \citet{hsiao21aaai}, open-sourced on GitHub.\footnote{\url{https://github.com/YatingMusic/compound-word-transformer}} It consists of 1,747 pure piano performances of various Japanese, Korean, and Western pop songs, amounting to a total duration of $\sim$100 hours. All the songs are in 4/4 time signature, namely four beats per bar (measure). We leave 5\% (87~songs) as the validation set.

According to \citet{hsiao21aaai}, the piano performances are originally collected from the Internet
in the MP3 (audio) format. \citeauthor{hsiao21aaai}\ further employed \textit{Onsets and Frames} piano transcription \cite{hawthorne2018onsets}, \texttt{madmom} beat tracking tool \cite{bock2016madmom}, and \texttt{chorder} rule-based chord detection\footnote{\url{https://github.com/joshuachang2311/chorder}} to transcribe the audio into MIDI format with tempo, beat, and chord information.

\vspace{-0.2cm}\paragraph*{Data representation.}
The representation adopted here is largely identical to the \textit{Revamped MIDI-derived} (REMI) encoding by \citet{huang2020pop}, except that an extended set of \texttt{chord} tokens (described below) is used. REMI encodes a piano piece into a sequence composed of two types, \textit{metrical} and \textit{note}, of tokens. The \textit{metrical} tokens are:
\begin{itemize}[leftmargin=*, noitemsep,topsep=0pt]
\item \texttt{bar}: Marks the start of a musical bar.
\item \texttt{subbeat}: Marks the musical timing within a bar. A bar is divided into $16$ \texttt{subbeat}s, which is equivalent to 4 beats. This symbolic timing provides an explicit time grid for sequence models to model music.
\item \texttt{tempo}: Determines the pace (in beats per minute, or \textit{bpm}) at which the piece is played, varied per bar. The range of \texttt{tempo} tokens is $[32, 224]$ bpm, in steps of $3$ bpm for quantization. 
\end{itemize}
The \textit{note} tokens are:
\begin{itemize}[leftmargin=*, noitemsep,topsep=0pt]
\item \texttt{pitch}: Marks a note played. The $88$ \texttt{pitch}-es correspond to each key on the piano.
\item \texttt{duration}: Denotes the length of a played note, ranging from $1/2$ to $16$ subbeats, in steps of $1/2$ subbeat.
\item \texttt{volume} (or, velocity): Denotes how loud a note is played. A total of $24$ \texttt{volume} levels are considered.
\item \texttt{chord}: Marks a change on the accompanying chord. Each \texttt{chord} is described by its \textit{root note} and \textit{quality}, e.g., \texttt{C-Maj7}, \texttt{E-min}. A total of $133$ distinct \texttt{chord} tokens are found in the dataset.
\end{itemize}
Please note that a single note played is represented by a co-occurring triple of (\texttt{pitch}, \texttt{duration}, \texttt{volume}). The aforementioned tokens constitute a vocabulary of size ${\sim}$340 for our REMI encoding.
On average, we need a sequence with 5\,300 tokens to represent a song.

\paragraph*{Training and inference.} In each training epoch, we randomly crop a segment of length 2\,048 from each sample, and shift the pitches of the entire segment by $-$6 to 6 semitones randomly (this is called \textit{transposition} in music) as data augmentation.
We use batch size $=4$, and set the learning rate to $0.0001$ for APE and $0.0002$ for all SPE models. For \texttt{sineSPE}, we choose the number of sines $K = 5$; for \texttt{convSPE}, the convolutional filter size is set to be $128$, $512$ for the gated and ungated variants respectively.

Detailed resource usage of each model is shown in Table \ref{tab:pop-resource}. 

During inference, we employ \textit{nucleus sampling} \cite{holtzman2019curious} with $p=0.9$ and softmax temperature $t=1.2$. No post-processing on enforcing the grammatical correctness of the generated sequence is done.

Validation loss of the models trained on this task is listed in Table  \ref{tab:remi-gen-perf}. On this metric, our \texttt{convSPE} variant performs the best both within the trained positions and on extrapolation.

\subsection{Groove Continuation}
\label{sec:groove-cont}
\paragraph*{Training data.}
The Groove2Groove MIDI dataset\footnote{\url{http://doi.org/10.5281/zenodo.3958000}} consists of accompaniments generated by the Band-in-a-Box software (BIAB).\footnote{\url{https://www.pgmusic.com/}}
We only use the training section of the Groove2Groove MIDI dataset and perform a custom training/validation/test split such that each section contains a unique set of BIAB styles (2\,761 for training and 50 each for validation and testing).
The code necessary to download, pre-process and split the dataset is included in the repository.

We convert each accompaniment to a trio consisting of bass, drums and another randomly selected accompaniment track (e.g.\ piano, guitar).
We then perform random data augmentation by skipping measures at the beginning, dropping some of the instruments, and transposition (pitch-shifting by $-$5 to $+5$ semitones).
All randomization is done anew in each epoch.

\paragraph*{Data representation.}
We use a representation similar to the one proposed by \citet{cifka:hal-02923548}, but adapted to a multi-track (multi-instrument) setting.
Specifically, we encode a piece of music as a sequence of the following types of 
event tokens, each with two integer arguments:
\begin{itemize}[leftmargin=*, noitemsep,topsep=0pt]
\item \texttt{note\_on(track, pitch)}: Begins a new note at the given pitch (0--127).
\item \texttt{note\_off(track, pitch)}: Ends the note at the given pitch (0--127).
\item \texttt{time\_shift(beats, offset)}: Advances current time by a given number of beats and then sets the offset within the beat, given as the number of ticks from its beginning (0--11). Maximum possible shift is \texttt{(2, 0)}.
\end{itemize}
The track numbers range from 1 to 3, where 1 is always bass and 2 is always drums. The vocabulary of the model then consists of 794 tokens (3$\times$128 note-ons, 3$\times$128 note-offs, 24 time shifts, and 2 beginning-/end-of-sequence markers).

The main differences to the representation described in Section \ref{sec:pop-generation} are a more compact encoding of timing, no representation of musical dynamics (for simplicity), and support for multiple tracks (not originally proposed by \citealp{cifka:hal-02923548} but introduced here inspired by \citealp{donahue2019lakhnes}).

\paragraph*{Training and inference.}
During training, each example is pre-processed and encoded as described above and the resulting token sequence is truncated to a length of 512.
We train each model for a total of 24 epochs.

At test time, we sample with a softmax temperature of 0.6.
We disallow sampling tokens that would result in invalid sequences (i.e.\ spurious note-offs, backward time shifts) in order to ensure that the generated sequence can be correctly decoded.

\paragraph*{Various training details.} Hyperparameter tuning was mostly performed in preliminary experiments ($\sim$100 runs); these were mostly done on other variants of the dataset and with different sequence lengths (ranging from 256 to 20\,k); this includes experiments discarded due to bugs discovered during or after training.
Learning rates between $0.0001$ and $0.0008$ and batch sizes between $1$ and $24$ were considered.
For SPE, we considered both the gated and ungated variants with as many realizations as fit in memory (between 16 and 64). 
Model selection was based on validation loss and informal perceptual evaluation.
Only a minimal attempt at further learning rate tuning was made for the final set of models with length 512, which did not appear to be particularly sensitive to it, and we chose to keep the initial learning rate $0.0004$, which was found to perform well in all cases.

The models included in the main document~-- \texttt{APE}, \texttt{sineSPE} and \texttt{convSPE}~-- all use a batch size of 10 and finished training in about 3\,h, 5\,h and 6\,h, respectively, using 9.7\,GB, 14.4\,GB and 14.8\,GB of GPU memory. The total training time including all preliminary experiments was 852 hours.

\paragraph*{Evaluation metrics.}
We use the objective metrics proposed by \citet{cifka-ismir2019,cifka:hal-02923548} to measure the style similarity between the generated continuation and the file from which the prompt was extracted.
Given two pieces of music, each metric gathers musical event statistics of the two pieces in histograms called \emph{style profiles}, and then computes the cosine similarity between them.

The two metrics used here, \emph{onset-duration} and \emph{time-pitch}, differ in what kind of events they use to construct the style profile:
\begin{itemize}[leftmargin=*, noitemsep,topsep=0pt]
    \item The \textbf{onset-duration} profile is defined as a 2D histogram relating note onset positions to note durations. More precisely, for all notes $a$ in a piece of music, it records a tuple of the form
    \begin{equation*}
    \begin{split}
    (\text{start}(a) \bmod 4,
    \> \text{end}(a)-\text{start}(a))
    \in [0,4)\times[0,2),
    \end{split}
    \end{equation*}
    where $\text{start}(a)$ and $\text{end}(a)$ refer to the onset and offset time of $a$ in beats.
    The expression $\text{start}(a) \bmod 4$ then represents the position of the note onset relative to the current bar, since all examples in the dataset are in a 4-beat meter.
    These tuples are gathered in $24\times12$ histogram bins (24 for onset time and 12 for duration).
    \item The \textbf{time-pitch} profile is also obtained as a 2D histogram, this time capturing time differences and pitch differences (intervals) between notes.
    The tuples it considers have the form
    \begin{equation*}
    \begin{split}
    (\text{start}(b)-\text{start}(a),
     \> \text{pitch}(b)-\text{pitch}(a)) \\
    \in [0,4)\times\{-20,-19,\ldots,20\},\ a\neq b,
    \end{split}
    \end{equation*}
    where $a,b$ is a pair of notes and $\text{pitch}(\cdot)$ represents the pitch of a note as its MIDI note number (the number of semitones from $\text{C}_{-1}$). The histogram has $24\times41$ bins (24 for time lags between 0 and 4 beats and 41 bins for intervals between $-20$ and $20$ semitones).
    
\end{itemize}
In both cases, the 2D histograms are flattened to vectors before computing cosine similarities.

\section{Additional Results}
\label{sec:additional}

\subsection{Attention Visualization: Music Generation}
\begin{figure*}
    \centering
    \begin{subfigure}[b]{0.46\linewidth}
        \centering
        \includegraphics[width=\textwidth]{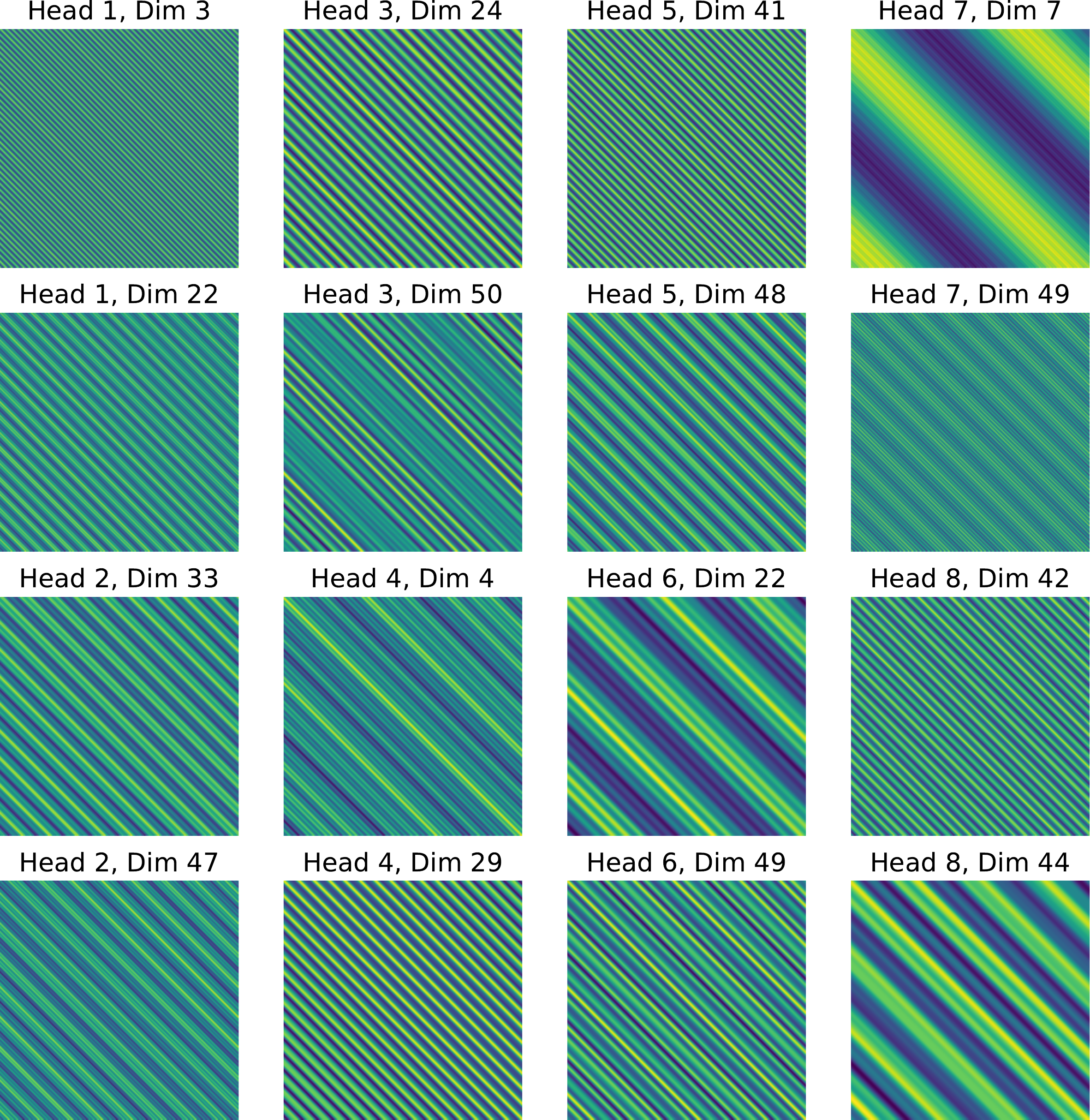}
        \caption{\texttt{sineSPE}}
        \label{subfig:sinespe-vis}
    \end{subfigure}
    \hspace{0.5cm}
    \begin{subfigure}[b]{0.46\linewidth}
        \centering
        \includegraphics[width=\textwidth]{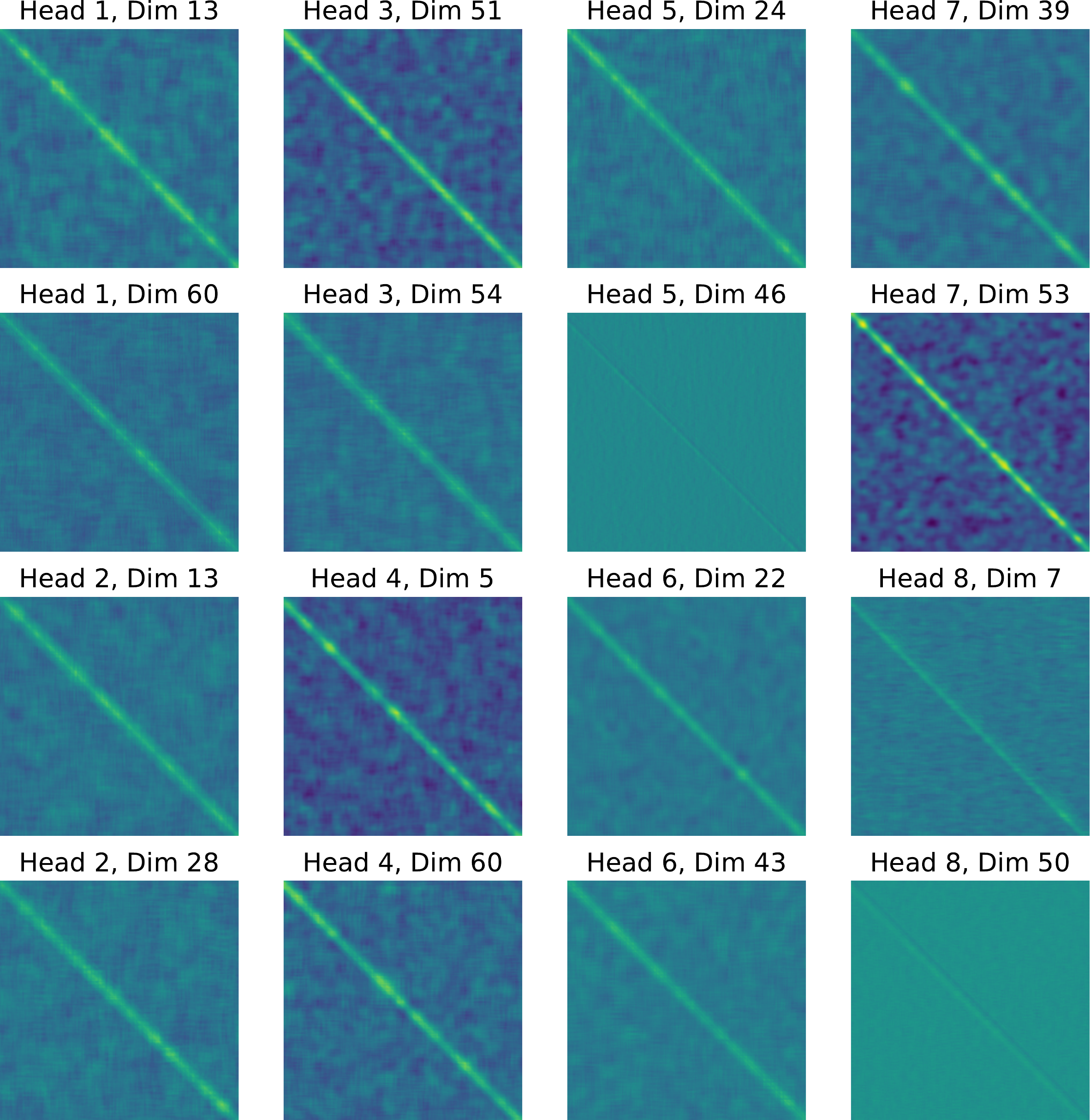}
        \caption{\texttt{convSPE}}
        \label{subfig:convspe-vis}
    \end{subfigure}
    \vspace{-0.2cm}
    \caption{Examples of $\textbf{P}_d$ learned by SPE. X- and Y-axes are \textit{key} and \textit{query} positions respectively. Max position  $=2\,048$.}
    \label{fig:spe-vis}
\end{figure*}

\begin{figure*}
    \centering
    \includegraphics[width=0.96\textwidth]{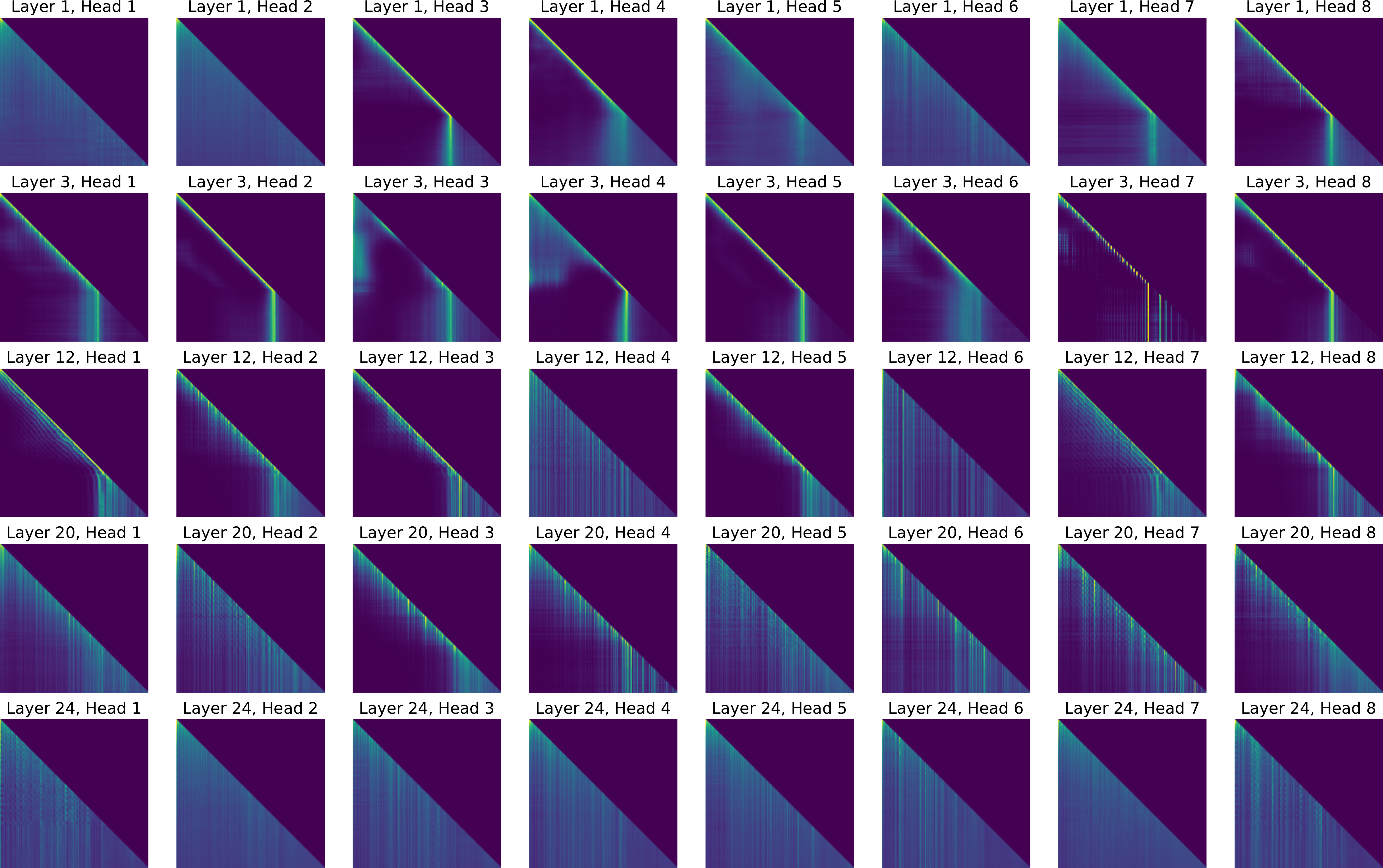}
    \caption{Full attention matrices of \texttt{APE} (baseline). X- and Y-axes are \textit{key} and \textit{query} positions respectively. Max position  $=3\,072$.}
    \label{fig:ape-attn}
\end{figure*}

\begin{figure*}
    \centering
    \includegraphics[width=0.96\textwidth]{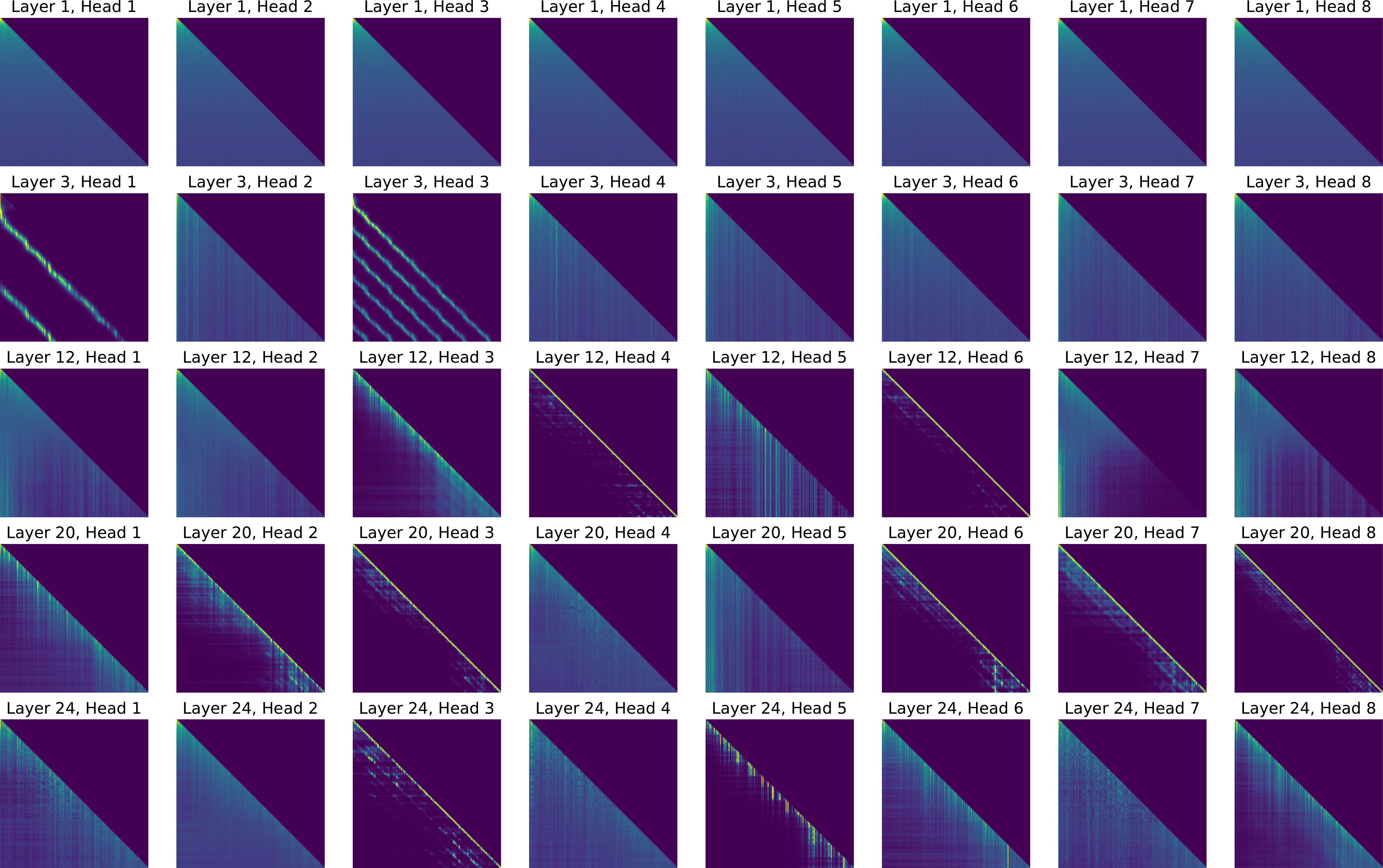}
    \vspace{-0.2cm}
    \caption{Full attention matrices of \texttt{sineSPE} (\textit{with} gated SPE). Max token position $=3\,072$.}
    \label{fig:sinespe-attn}
\end{figure*}

\begin{figure*}
    \centering
    \includegraphics[width=0.96\textwidth]{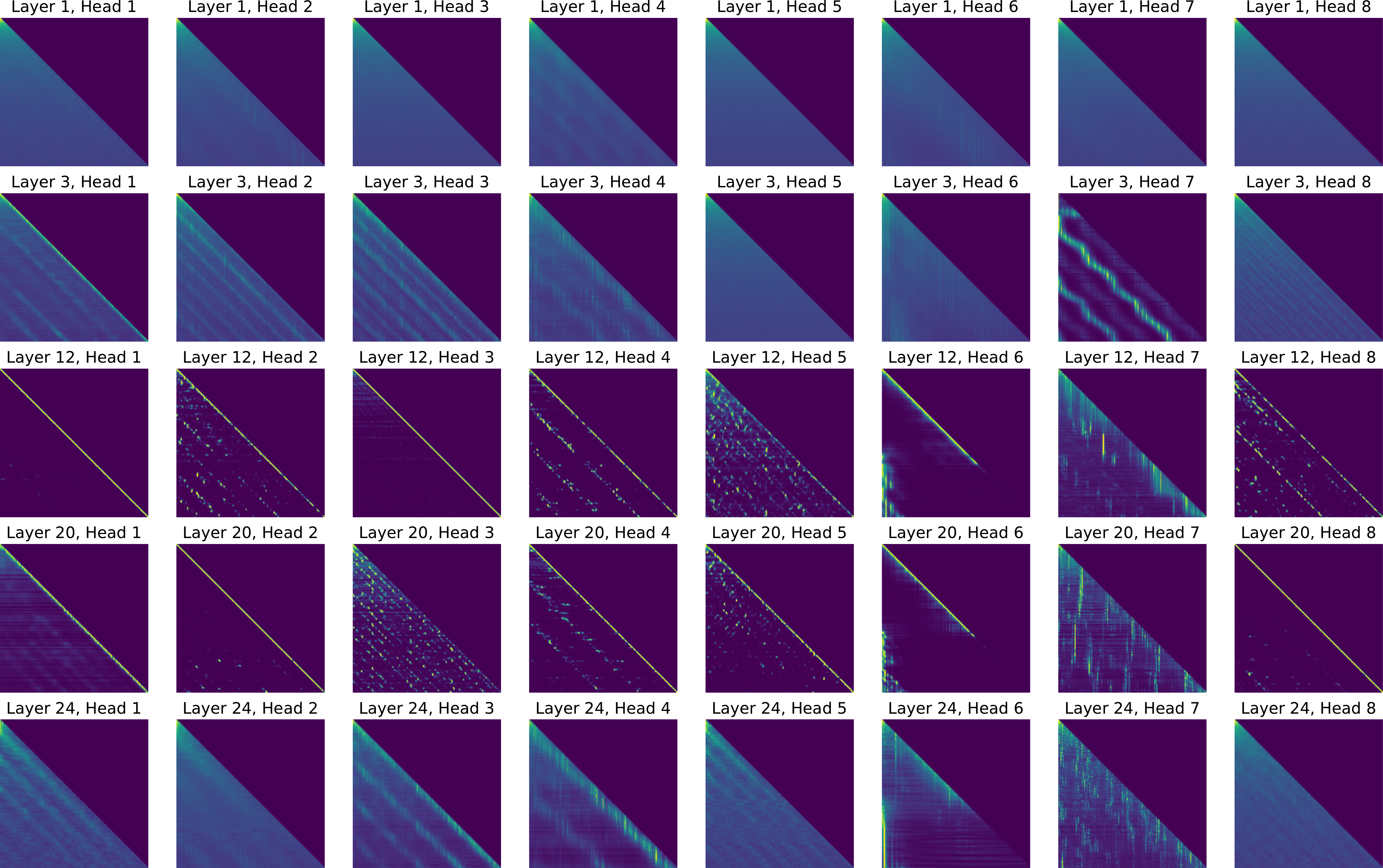}
    \vspace{-0.2cm}
    \caption{Full attention matrices of \texttt{sineSPE} (\textit{without} SPE gating). Max token position $=3\,072$.}
    \label{fig:sinespe-ug-attn}
\end{figure*}

\begin{figure*}
    \centering
    \includegraphics[width=0.96\textwidth]{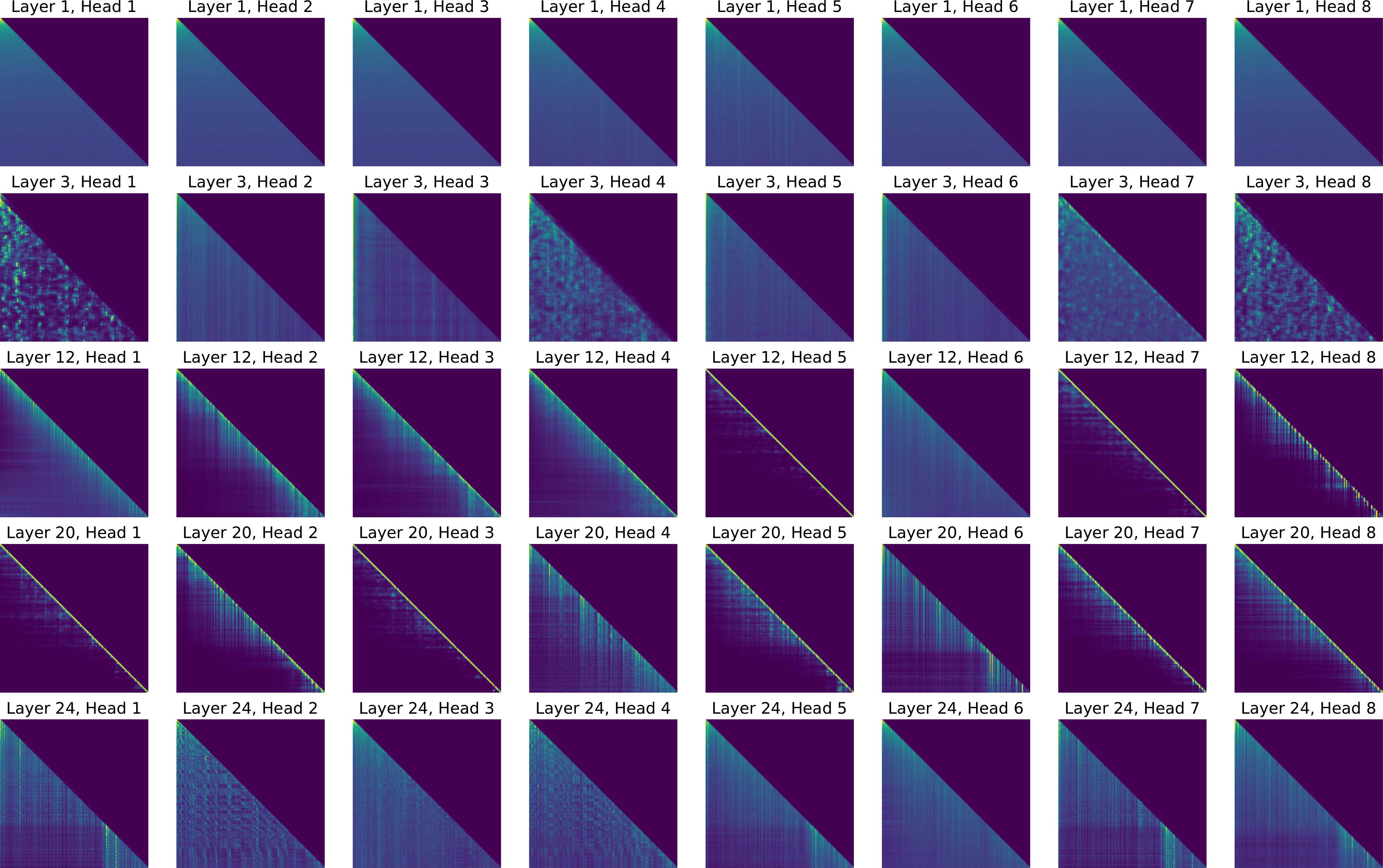}
    \vspace{-0.2cm}
    \caption{Full attention matrices of \texttt{convSPE} (\textit{with} SPE gating, conv filter size $=128$). Max token position $=3\,072$.}
    \label{fig:convspe-attn}
\end{figure*}

\begin{figure*}
    \centering
    \includegraphics[width=0.96\textwidth]{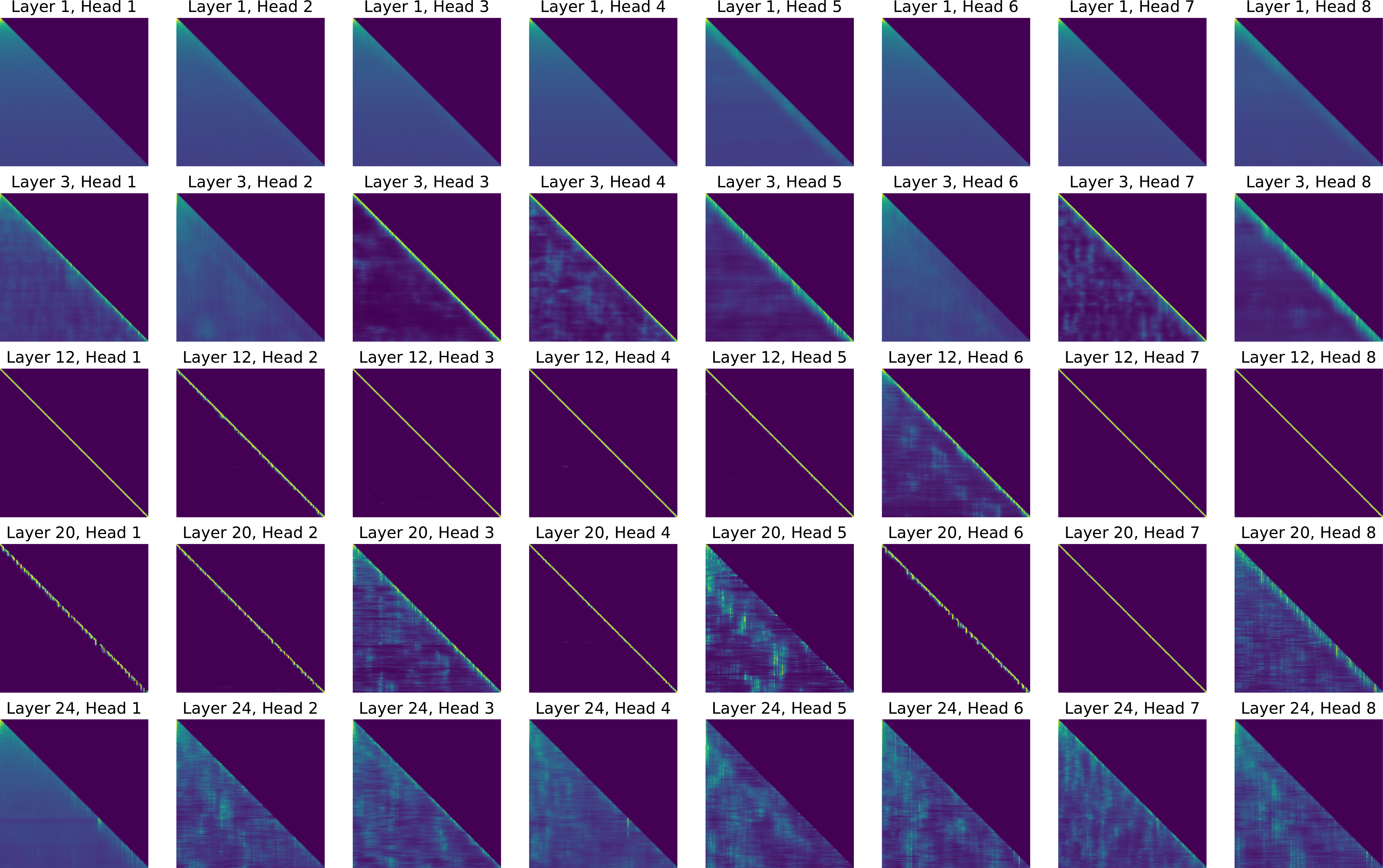}
    \vspace{-0.2cm}
    \caption{Full attention matrices of \texttt{convSPE} (\textit{without} SPE gating, conv filter size $=512$). Max token position $=3\,072$.}
    \label{fig:convspe-ug-attn}
\end{figure*}

In this section, we display attention patterns produced by our pop piano music generation models.

\paragraph*{Learned positional templates.}
We share the SPE modules across all layers of the Performer, but not across the attention heads, resulting in $512$ learned positional kernels $\mathcal{P}_d$) (\textit{number of heads} $\times$ \textit{key dimensions per head}. In Figure \ref{fig:spe-vis}, we display 16 randomly picked resulting templates $\textbf{P}_d$ for both \texttt{sineSPE} and \texttt{convSPE}, trained with gating. Details of the two variants are:
\begin{itemize}[leftmargin=*, noitemsep,topsep=0pt]
    \item \texttt{sineSPE}: We set the number of sines $K=5$.
    \item \texttt{convSPE}: We use filters of size $128$.
\end{itemize}
In accordance with the definition, all of the visualizations are plotted with the equation $\textbf{P}_d~=~ \overline{\textbf{Q}}_d \overline{\textbf{K}}_d^\top$, which we never need to explicitly compute for linear transformers. From Figure \ref{fig:spe-vis}, we can observe that \texttt{sineSPE} learns to exploit a wide range of frequencies, and that \texttt{convSPE} is effective within small query-key offsets corresponding to the filter size, as expected.

\paragraph*{Full Attention.}
Although the full attention matrix $\mathbf{A}$ is not computed in linear transformers, we can still obtain it \textit{offline} by multiplying queries and keys through either $\textbf{A}~=~\exp(\textbf{Q}\textbf{K}^{\top}/\sqrt{D})$ (in the case of APE, where $D$ is the key dimensions per head), or $\textbf{A}=\exp(\widehat{\textbf{Q}}\widehat{\textbf{K}}^\top/\sqrt{R})$ (in the case of SPEs); then apply row-wise softmax operation on $\textbf{A}$ as normalization.

Here, we present the (softmax-ed) attention matrices in the $1$st, $3$rd, $12$th, $20$th, and $24$th (last) layers of all the five models trained on pop piano music generation in Figures \ref{fig:ape-attn}--\ref{fig:convspe-ug-attn}. These are computed from one of each model's random from-scratch music generations. 
To examine the models' extrapolation ability, we let them generate a sequence of length $3\,072$, while the training sequence length is only $2\,048$. The attention matrices are lower-triangular due to causal masking. For better visualization, the color of each pixel is adjusted through $\min\{1, {a_{mn}}^{0.4} / 0.02^{0.4}\}$ in the plots, where $a_{mn} \in [0, 1]$ is the softmax-ed attention score.

Figure \ref{fig:ape-attn} reveals a major drawback of \texttt{APE}: the attention of tokens beyond position 2\,048 (the training sequence length) seems to concentrate around 2\,048 in earlier layers, rather than paying global or local attention. 
Such behavior is not seen in any of our SPE models.
This potentially explains \texttt{APE}'s poor generalization to long sequences suggested by the stark increase in validation loss after position 2\,048 (see Figure 3 in the main paper, and Table \ref{tab:remi-gen-perf} here).

Next, comparing Figures \ref{fig:sinespe-attn} and \ref{fig:sinespe-ug-attn}, it is obvious that gated SPE gives the model the freedom to \textit{switch off} PE in some heads to achieve global attention (see Figure \ref{fig:sinespe-attn}), whereas the attention of ungated \texttt{sineSPE} (Figure \ref{fig:sinespe-ug-attn}) largely stays periodic, which might not be always desirable. The same can be said for \texttt{convSPE} (Figures \ref{fig:convspe-attn} and \ref{fig:convspe-ug-attn}). The gated \texttt{convSPE} is able to look much further back in the middle layers than its ungated counterpart.

\subsection{Attention Visualization: CIFAR10}
\begin{figure*}
    \centering
    \includegraphics[width=\linewidth]{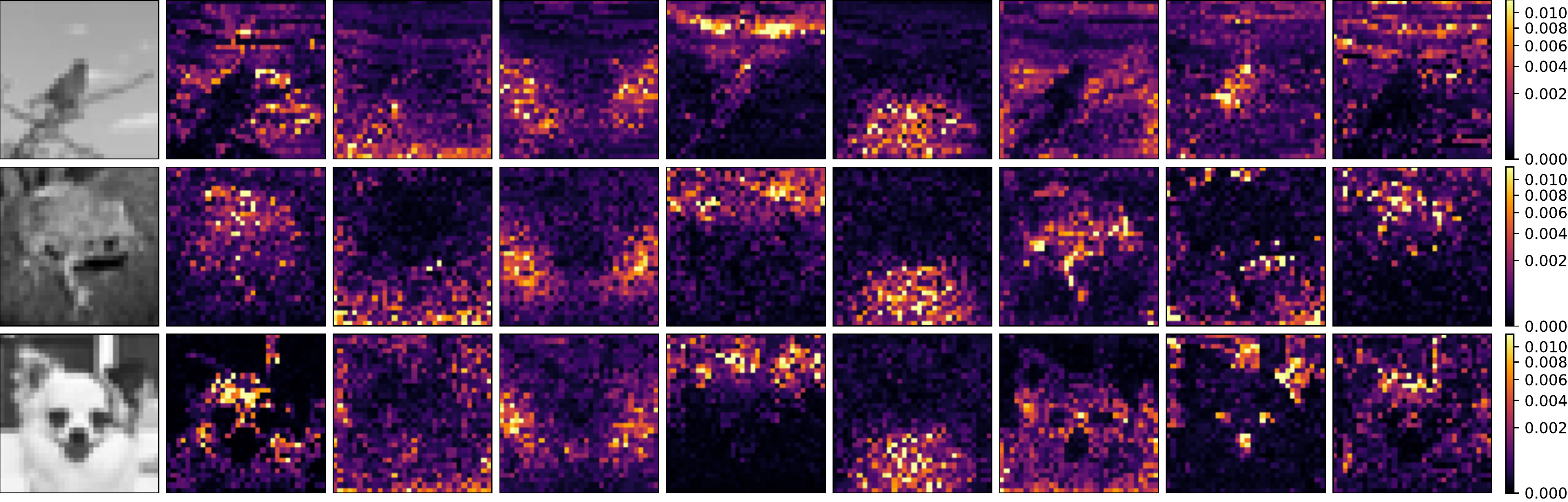}\\[.25cm]
    \includegraphics[width=\linewidth]{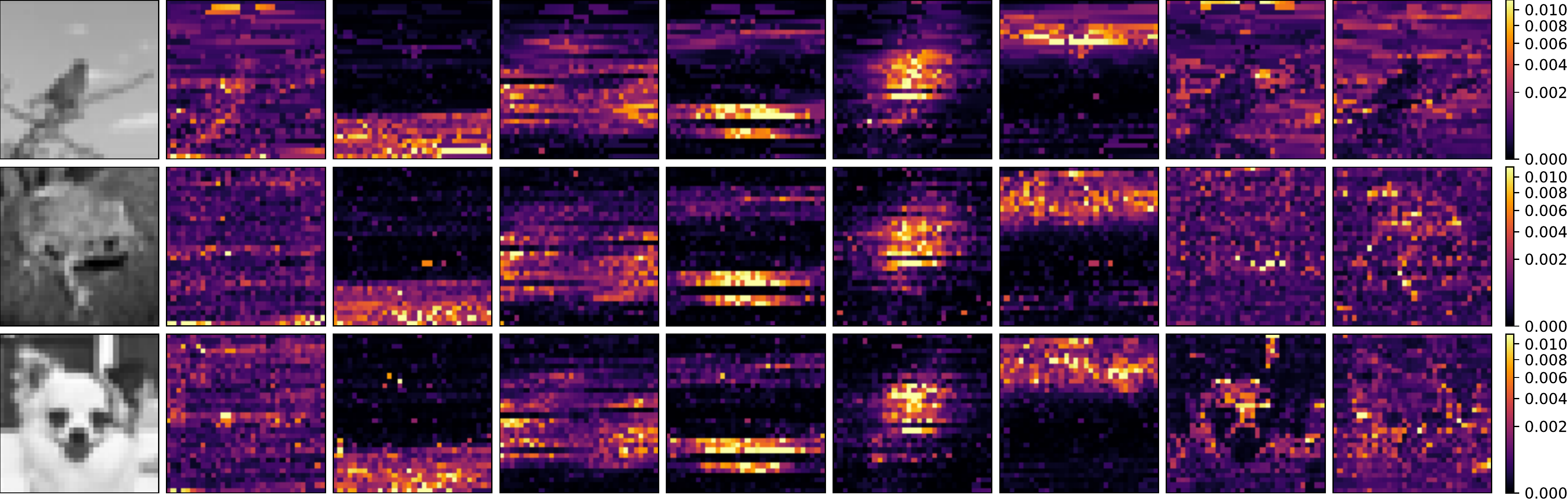}\\[.25cm]
    \includegraphics[width=\linewidth]{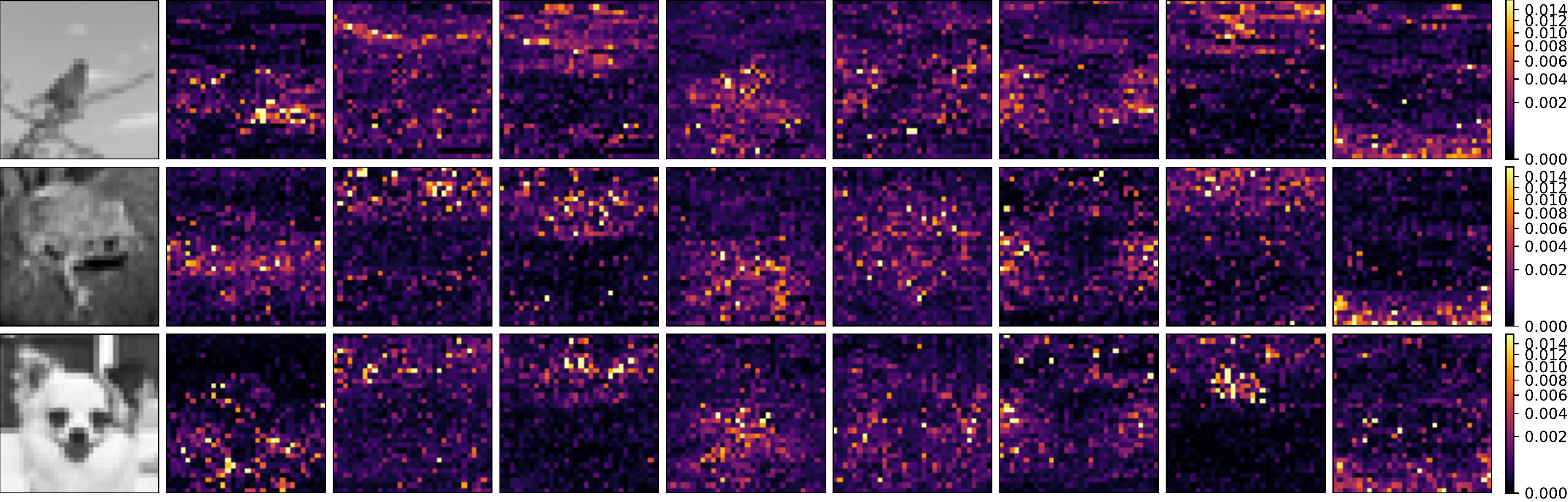}
    \caption{CIFAR10 attention maps for 3 variants of Linear Transformer-ReLU: learnable APE (top), \texttt{sineSPE} (middle), and \texttt{convSPE} (bottom). Each row displays the input image, followed by attention weights of the 8 respective heads for each pixel, with the special \texttt{CLS} token as the query.}
    \label{fig:cifar10_attn}
\end{figure*}

Figure \ref{fig:cifar10_attn} displays attention maps extracted from models trained on the LRA CIFAR10 task.
Note that these are one-layer networks, and classification is done by prepending a special \texttt{CLS} token to the sequence of pixel values and using the output at this first position as input to a feed-forward classifier.
Consequently, only the attention map at this single position (which is the one we display here) matters.
(The model is therefore \emph{de facto} not using self-attention, but rather attention with a single query and many keys.
This removes the distinction between relative and absolute positions, which might explain why trainable APE performs better than SPE on this task.)

\subsection{Evaluation of Desired PE Properties}

\begin{table*}
    \caption{Evaluation of PEs metrics. \texttt{T:} translation invariance, \texttt{M:} monotonicity (lower is better). \texttt{ug}: models trained without SPE gating.}
    \vskip 0.1in  %
    \label{tab:pe-props}
    \scriptsize
    \renewcommand{\arraystretch}{1.25}
    \centering
    \begin{tabularx}{\linewidth}{lCCC|CCC?CCC}
    \toprule
     \multicolumn{1}{r}{\textbf{Query positions}} & \multicolumn{3}{c|}{$\mathbf{0 < \textbf{pos} \leq 1\,024}$} & \multicolumn{3}{c?}{$\mathbf{1\,024 < \textbf{pos} \leq 2\,048}$} & \multicolumn{3}{c}{$\mathbf{2\,048 < \textbf{pos} \leq 2\,560}$ (extrapolation)} \\
      \multicolumn{1}{r}{\textit{Query-key offset}} & ${<}128$ & ${<}512$ & ${<}1\,024$ & ${<}128$ & ${<}512$ & ${<}1\,024$ & ${<}128$ & ${<}512$ & ${<}1\,024$\\
    \midrule
    \texttt{APE} & \mccell{\texttt{T:} 0.4335\\ \texttt{M:} \textbf{0.0152}} & \mccell{\texttt{T:} 0.2063\\ \texttt{M:} 0.0625} & \mccell{\texttt{T:} 0.1845\\ \texttt{M:} \textbf{0.0616}} & \mccell{\texttt{T:} 0.9142\\ \texttt{M:} \textbf{0.0193}} & \mccell{\texttt{T:} 0.6953\\ \texttt{M:} 0.0413} & \mccell{\texttt{T:} 0.6458\\ \texttt{M:} \textbf{0.0713}} & \mccell{\texttt{T:} 0.9599\\ \texttt{M:} 0.3974} & \mccell{\texttt{T:} 0.8959\\ \texttt{M:} 0.2429} & \mccell{\texttt{T:} 0.5886\\ \texttt{M:} 0.1637} \\ \hline
    \texttt{sineSPE} & \mccell{\texttt{T:} 0.1660\\ \texttt{M:} 0.2893} & \mccell{\texttt{T:} 0.3078\\ \texttt{M:} 0.4406} & \mccell{\texttt{T:} 0.3527\\ \texttt{M:} 0.4283} & \mccell{\texttt{T:} 0.1337\\ \texttt{M:} 0.2826} & \mccell{\texttt{T:} 0.2504\\ \texttt{M:} 0.4063} & \mccell{\texttt{T:} 0.3228\\ \texttt{M:} 0.4167} & \mccell{\texttt{T:} 0.2167\\ \texttt{M:} 0.3253} & \mccell{\texttt{T:} 0.3599\\ \texttt{M:} 0.4060} & \mccell{\texttt{T:} 0.4147\\ \texttt{M:} 0.3913} \\ \hline
    \texttt{sineSPE} (\texttt{ug}) & \mccell{\texttt{T:} \textbf{0.0141}\\ \texttt{M:} 0.6295} & \mccell{\texttt{T:} 0.0242\\ \texttt{M:} 0.1844} & \mccell{\texttt{T:} 0.0231\\ \texttt{M:} 0.1582} & \mccell{\texttt{T:} \textbf{0.0135}\\ \texttt{M:} 0.6238} & \mccell{\texttt{T:} 0.0206\\ \texttt{M:} 0.1623} & \mccell{\texttt{T:} 0.0190\\ \texttt{M:} 0.1061} & \mccell{\texttt{T:} \textbf{0.0105}\\ \texttt{M:} 0.6189} & \mccell{\texttt{T:} 0.0196\\ \texttt{M:} 0.1609} & \mccell{\texttt{T:} 0.0163\\ \texttt{M:} 0.0994} \\ \hline
    \texttt{convSPE} & \mccell{\texttt{T:} 0.3422\\ \texttt{M:} 0.1781} & \mccell{\texttt{T:} 0.5637\\ \texttt{M:} 0.2242} & \mccell{\texttt{T:} 0.6389\\ \texttt{M:} 0.2189} & \mccell{\texttt{T:} 0.3209\\ \texttt{M:} 0.1735} & \mccell{\texttt{T:} 0.6239\\ \texttt{M:} 0.3624} & \mccell{\texttt{T:} 0.7648\\ \texttt{M:} 0.4192} & \mccell{\texttt{T:} 0.3462\\ \texttt{M:} 0.1486} & \mccell{\texttt{T:} 0.6135\\ \texttt{M:} 0.3247} & \mccell{\texttt{T:} 0.7025\\ \texttt{M:} 0.2740} \\ \hline
    \texttt{convSPE} (\texttt{ug}) & \mccell{\texttt{T:} 0.2828\\ \texttt{M:} 0.1234} & \mccell{\texttt{T:} \textbf{0.0192}\\ \texttt{M:} \textbf{0.0249}} & \mccell{\texttt{T:} \textbf{0.0107}\\ \texttt{M:} 0.0620} & \mccell{\texttt{T:} 0.3334\\ \texttt{M:} 0.1505} & \mccell{\texttt{T:} \textbf{0.0188}\\ \texttt{M:} \textbf{0.0253}} & \mccell{\texttt{T:} \textbf{0.0109}\\ \texttt{M:} 0.1254} & \mccell{\texttt{T:} 0.2207\\ \texttt{M:} \textbf{0.1342}} & \mccell{\texttt{T:} \textbf{0.0171}\\ \texttt{M:} \textbf{0.0217}} & \mccell{\texttt{T:} \textbf{0.0106}\\ \texttt{M:} \textbf{0.0989}} \\
    \bottomrule
    \end{tabularx}
\end{table*}

We employ \textit{identical word probing} and the associated metrics introduced in \citet{wang2021on} to compare the \textit{translation invariance} and \textit{monotonicity}  properties of APE and our SPEs. The other properties mentioned in that work, namely \textit{symmetry} and \textit{direction balance}, are not evaluated here since the attention is uni-directional in our case. The models are also trained on pop piano music generation.

The metrics are calculated from attention matrices of each head in the $1$st layer, averaged over all possible \textit{identical-token} sequences (i.e., a sequence composed of repeated, same tokens; there are $\sim$340 of them for our REMI vocabulary). 
To eliminate the impact of applying row-wise softmax with causal masking on the translation invariance property, we compute the metrics on the \textit{unnormalized} attention matrices, %
i.e., $\textbf{A}=\exp(\textbf{Q}\textbf{K}^{\top}/\sqrt{D})$ for APE, and $\textbf{A}=\exp(\widehat{\textbf{Q}}\widehat{\textbf{K}}^\top/\sqrt{R})$ for SPEs. 
Various combinations of \textit{query positions} and \textit{query-key offsets} are considered to examine whether the PE properties stay consistent when we extrapolate to longer sequences, as well as to look into their behavior in local and long-range attention spans.

We report the scores of the best-performing (i.e., lowest-scoring) head of each model in Table \ref{tab:pe-props}. From the table, we can notice that the PE properties of \texttt{APE} often deteriorate drastically in cases of extrapolation. On the contrary, the scores of ungated SPE models, i.e., models in which we enforce the incorporation of positional information in every layer, remain remarkably consistent throughout the positions. The evaluation here provides additional evidence for the extrapolation capability of SPEs.

\subsection{Impact of the Number $R$ of Realizations}

\begin{figure*}[h]
    \centering
    \begin{subfigure}[b]{0.46\linewidth}
        \centering
        \includegraphics[width=\textwidth]{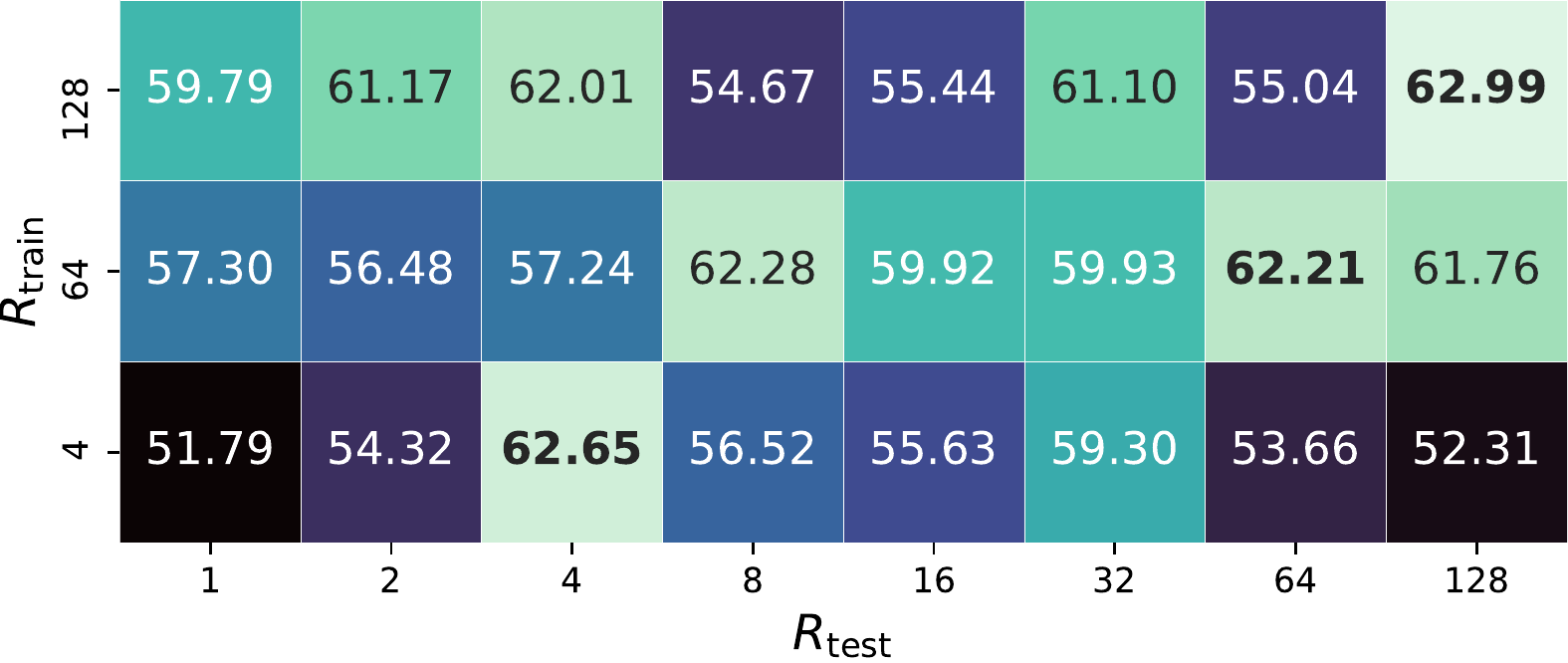}
        \caption{\texttt{sineSPE}}
    \end{subfigure}
    \hspace{0.5cm}
    \begin{subfigure}[b]{0.46\linewidth}
        \centering
        \includegraphics[width=\textwidth]{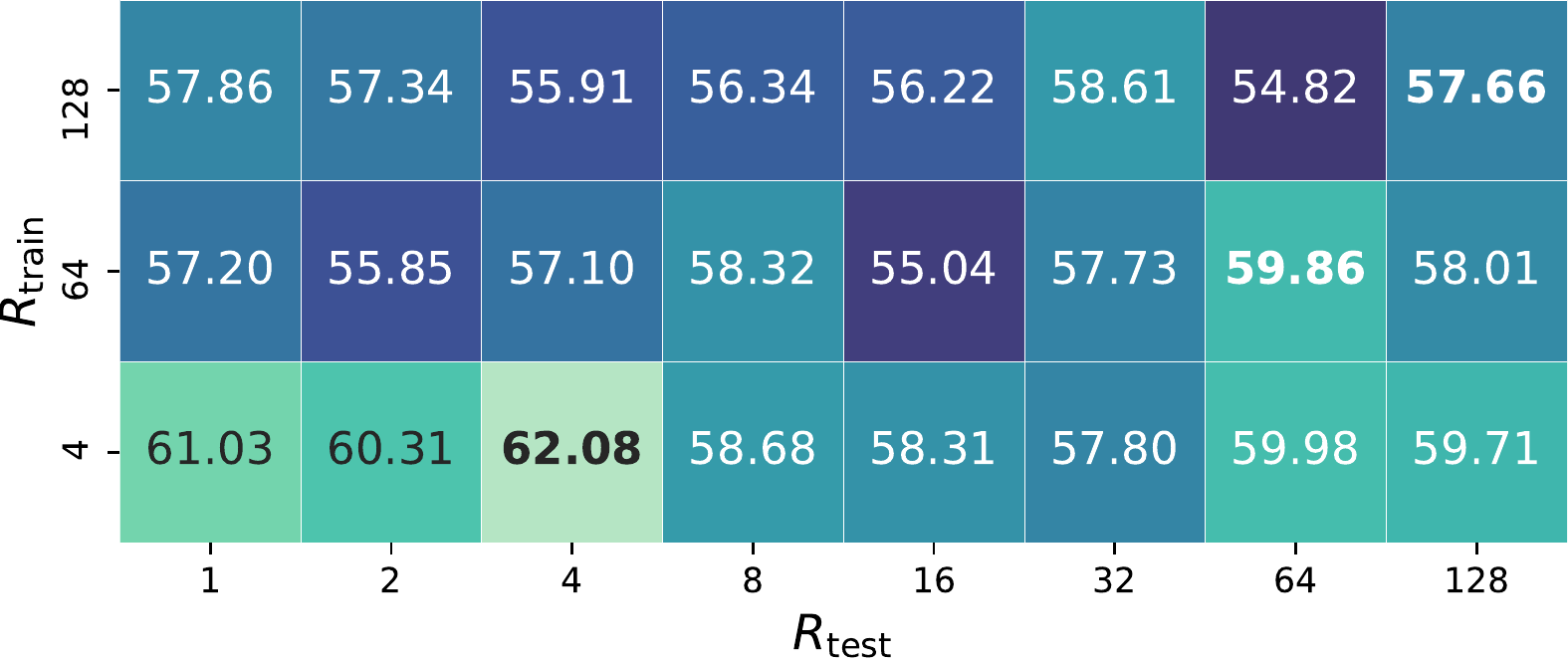}
        \caption{\texttt{convSPE}}
    \end{subfigure}
    \vspace{-0.2cm}
    \caption{Accuracy of Performer-softmax with SPE on the LRA Text task, with different numbers of realizations $R$ during training/testing. Each value is the result of a single run. Highlighted in bold are values obtained with $R_\text{test}=R_\text{train}$. Higher (brighter) is better.}
    \label{fig:ablation_R}
\end{figure*}

In the main document, we discussed how SPE asymptotically leads to the desired cross-covariance structure as $R$ grows to infinity. In this section, we empirically study how performance is affected by that parameter in practice. A first thing to highlight is that each training batch yields a new set of realizations for the noise $\textbf{Z}_d$, so that the network sees the right attention pattern \textit{on average}.

However, we may wonder whether how the number of realizations $R$ impacts training and test performance. One can indeed notice that $R$ may totally be set differently during training and inference, since it has no impact on the shape of the actual parameters/structure of the model. For this reason, we performed an ablation study where we use different values for $R_\text{train}$ at training time, resulting in a trained model, and then evaluate its performance using a possibly different value $R_\text{test}$. The results are displayed in Figure~\ref{fig:ablation_R}.

We can notice that the result achieved with $R_\text{test}=R_\text{train}$ (highlighted in bold) is consistently close to the best result for the same $R_\text{train}$, and conversely,
choosing $R_\text{test}\neq R_\text{train}$ often leads to a poor result.
In other words, training and testing with the same $R$ appears to be favorable for consistently good performance.

Another remarkable fact is that a higher $R$ does not seem to imply better performance, even when $R_\text{test}=R_\text{train}$. On the contrary, \texttt{convSPE} achieved by far the highest accuracy with $R=4$. This unexpected result seems contradictory to the fact that it means noisier attention patterns. Further investigation is required to explain this phenomenon, but we conjecture that this additional noise in the attention patterns leads to increased robustness of the trained model, helping generalization.

\end{document}